\documentclass[final,3p,times,twocolumn]{elsarticle}

\usepackage{amssymb}
\usepackage{amsmath}

\usepackage{xspace}
\usepackage{multirow}
\usepackage[table,xcdraw]{xcolor}
\usepackage[linesnumbered,ruled,vlined]{algorithm2e}
\usepackage{graphicx}
\usepackage[section]{placeins}
\usepackage{float}
\usepackage{amsmath}
\usepackage{soul}
\usepackage{subfig}
\usepackage{booktabs}
\usepackage{adjustbox}

\begin{document}

\sloppy

\begin{frontmatter}



\title{Reward Guidance for Reinforcement Learning Tasks Based on Large Language Models: The LMGT Framework\\
{\footnotesize \textsuperscript{*}This article has been accepted for publication in Knowledge-Based Systems. \copyright~Elsevier. Personal use is permitted, but republication/redistribution requires Elsevier permission.}}





\author[1]{Yongxin Deng\fnref{fn1}}
\author[1]{Xihe Qiu\fnref{fn1}\corref{cor1}}
\author[1]{Jue Chen\fnref{fn1}}
\author[2]{Xiaoyu Tan\fnref{fn1}}

\fntext[fn1]{These authors contributed equally to this work.}
\cortext[cor1]{Corresponding author. Email: qiuxihe@sues.edu.cn}

\affiliation[1]{organization={School of Electronic and Electrical Engineering, Shanghai University of Engineering Science},
            city={Shanghai},
            country={China}}

\affiliation[2]{organization={INFLY TECH (Shanghai) Co., Ltd.},
            city={Shanghai},
            country={China}}






\begin{abstract}
The inherent uncertainty in the environmental transition model of Reinforcement Learning (RL) necessitates a delicate balance between exploration and exploitation. This balance is crucial for optimizing computational resources to accurately estimate expected rewards for the agent. In scenarios with sparse rewards, such as robotic control systems, achieving this balance is particularly challenging. However, given that many environments possess extensive prior knowledge, learning from the ground up in such contexts may be redundant. To address this issue, we propose \textbf{L}anguage \textbf{M}odel \textbf{G}uided reward \textbf{T}uning (\textbf{LMGT}), a novel, sample-efficient framework. LMGT leverages the comprehensive prior knowledge embedded in Large Language Models (LLMs) and their proficiency in processing non-standard data forms, such as wiki tutorials. By utilizing LLM-guided reward shifts, LMGT adeptly balances exploration and exploitation, thereby guiding the agent's exploratory behavior and enhancing sample efficiency. We have rigorously evaluated LMGT across various RL tasks and evaluated it in the embodied robotic environment Housekeep. Our results demonstrate that LMGT consistently outperforms baseline methods. Furthermore, the findings suggest that our framework can substantially reduce the computational resources required during the RL training phase.
\end{abstract}



\begin{keyword}
Reinforcement Learning \sep Deep Learning Methods \sep Learning from Demonstration


\end{keyword}

\end{frontmatter}



\section{Introduction}

Reinforcement Learning (RL) faces a fundamental challenge in striking an optimal balance between exploration and exploitation \cite{yogeswaran2012reinforcement, fruit2019exploration}. This equilibrium is crucial for ensuring the robustness of RL algorithms in real-world applications \cite{busoniu2008comprehensive}. Agents in these environments encounter the exploration-exploitation dilemma due to the inherent unknown and stochastic nature of their surroundings, making it impossible to deduce the exact environmental model.

\begin{figure}[htbp]  
    \centering
    \subfloat[Captioner.]{
        \includegraphics[width=0.48\textwidth]{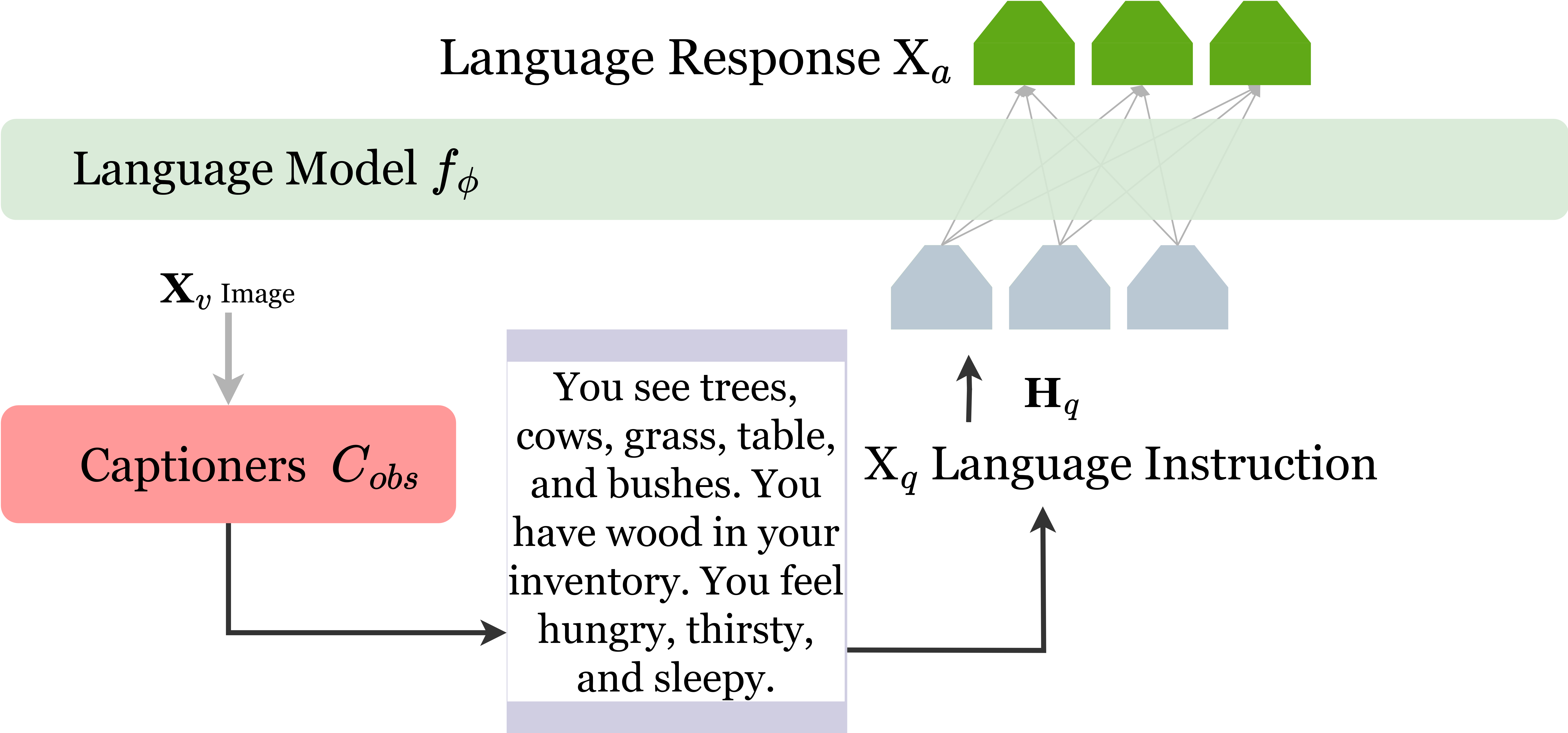}
    }\\ 
    \subfloat[End to End.]{
        \includegraphics[width=0.48\textwidth]{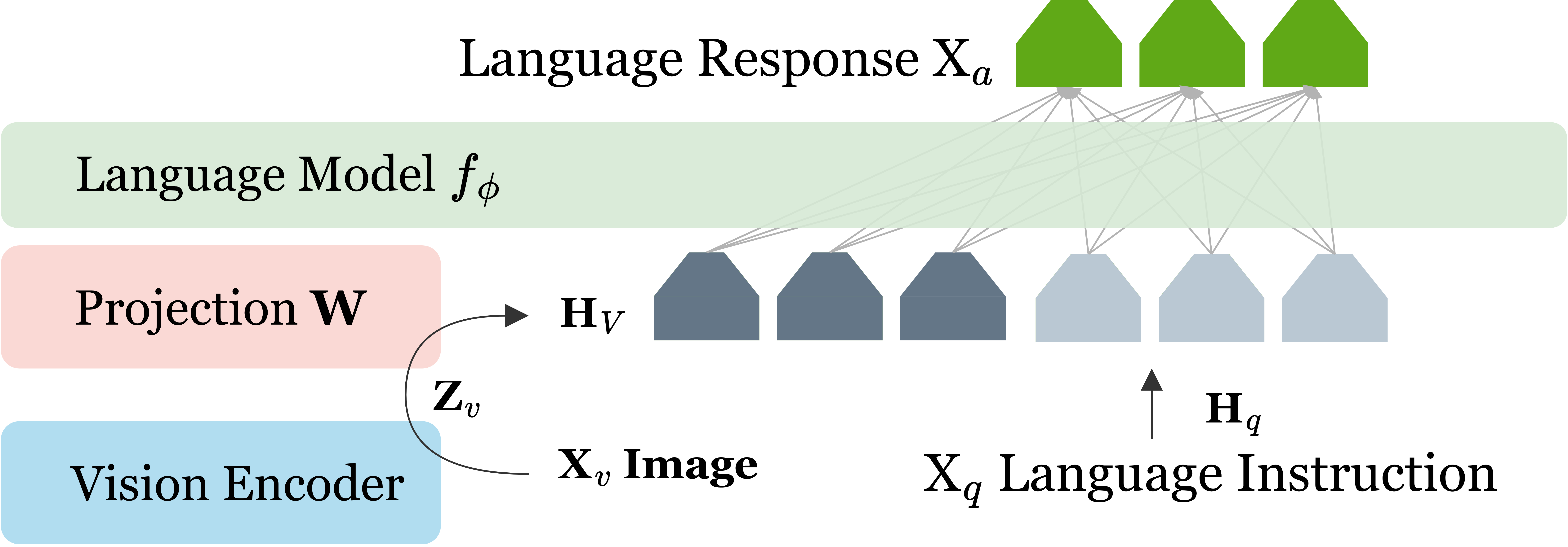}
    }
    \caption{\textbf{Schematic Representation of Diverse Approaches to Processing Environmental Information.} It is evident that leveraging Visual Instruction Tuning in an end-to-end framework significantly enhances the capacity of LLMs to assimilate more pertinent information for informed decision-making, compared to the captioner-based approach.}  
    \label{fig:llava}
\end{figure}

Through interactions with the environment, agents obtain estimates of expected rewards, denoted as $\hat{E}(R)$, for various actions. However, estimating rewards using sample statistics inherently introduces error, preventing certainty that the action with the highest $\hat{E}(R)$ is truly optimal. This uncertainty necessitates exploration to reduce the discrepancy between sample means and true means. Conversely, in resource-constrained scenarios, it is often advantageous to select actions with the highest estimated rewards—a practice known as exploitation. The accurate estimation of optimal actions is fundamental to RL \cite{kaelbling1996reinforcement, chen2022model}, whereas suboptimal actions require less computational evaluation. This inherent conflict necessitates selecting actions that balance proximity to the current ``best action'' estimate while maintaining sufficient diversity \cite{sutton2018reinforcement}.

Numerous approaches have attempted to address this challenge. Traditional strategies include $\epsilon$-greedy \cite{sutton2018reinforcement}, Softmax \cite{sutton2018reinforcement}, upper confidence bound \cite{jin2018q}, and Thompson sampling \cite{russo2018tutorial}. The selection of an appropriate exploration-exploitation strategy typically depends on the specific application context and problem requirements, as different RL problems demand different approaches. However, with the expanding scope of RL applications, manually selecting distinct strategies for each environment becomes impractical. Particular difficulties arise from multimodal and long-tailed data distributions. While some adaptive algorithms adjust the exploration-exploitation balance based on agent experiences \cite{zhang2021noveld}, these algorithms still have constraints that can significantly impact model performance and robustness when applied beyond their intended scope. Moreover, previous strategies either rely solely on adjusting ratios based on data distribution without utilizing prior knowledge, or require substantial domain expertise and task understanding, potentially reducing learning performance if improperly designed.

To overcome these limitations, we propose a novel framework called \textbf{L}anguage \textbf{M}odel \textbf{G}uided reward \textbf{T}uning (i.e., \textbf{LMGT}), which leverages prior knowledge from various sources to guide agents' learning with limited resources. Based on the work of Bruce \textit{et al.} \cite{bruce2022learning}, who demonstrated that valuable information can be derived from effective offline demonstration data, our approach enables agents to align themselves correctly with desirable behaviors. By capturing patterns of sound policies and using these for intrinsic motivation, RL agents can effectively map themselves to skillful demonstration-defined subspaces where even undirected exploration can significantly enhance environmental understanding if deviations align with rational pathways.

\begin{figure*}[htbp]
  \centering
  \includegraphics[width=0.85\linewidth]{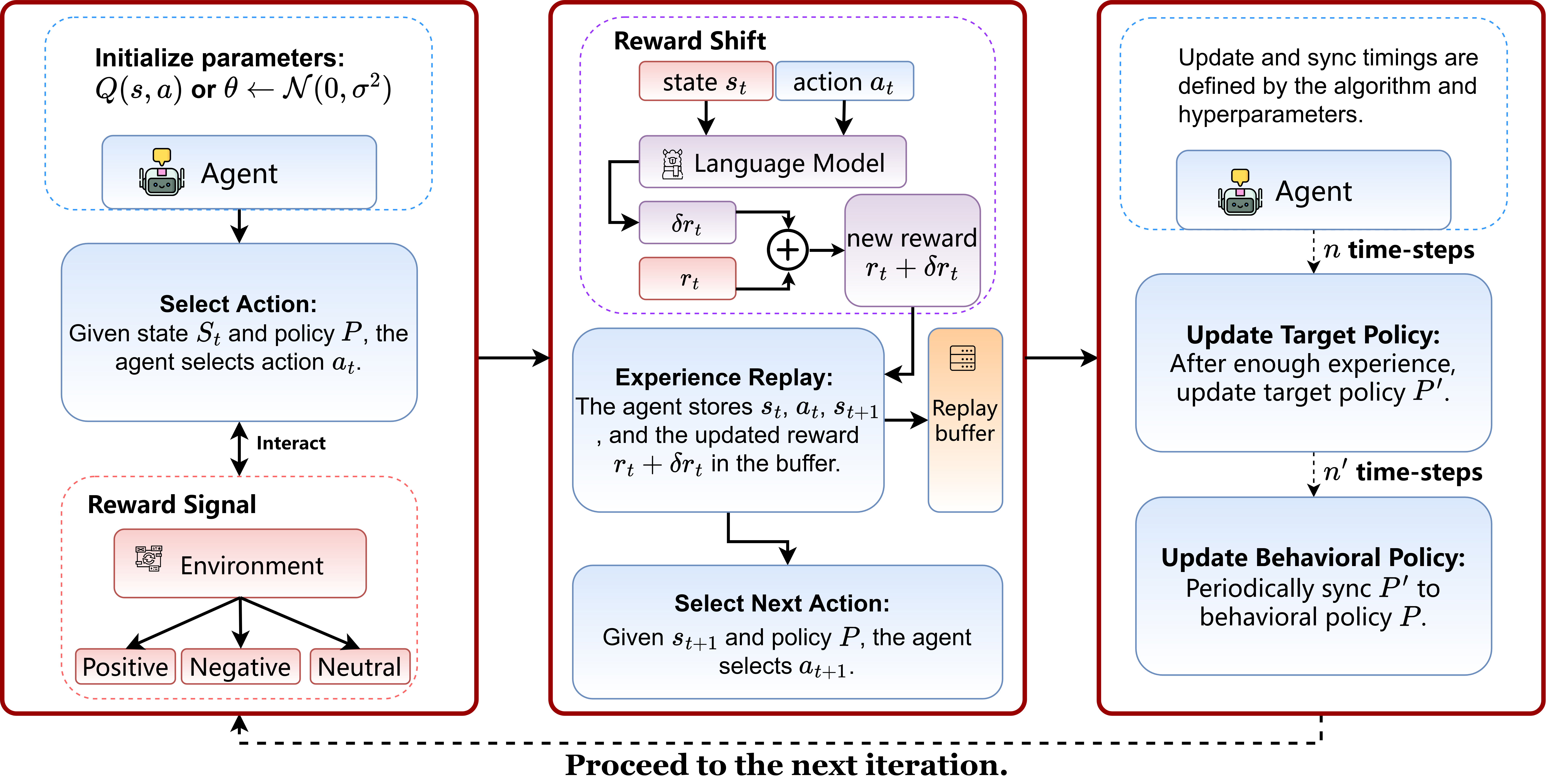}
  \caption{\textbf{The structure of our LMGT framework.} The LLM can observe the environment's state and the actions selected by the agent. It will evaluate the agent's behavior using prior knowledge, adjusting the final reward accordingly (via reward shifting). Thus, the agent’s stored experience inherently includes a component of prior knowledge.}
  \label{fig_lmgt}
\end{figure*}

Our LMGT framework harnesses the text comprehension and generation capabilities of Large Language Models (LLMs) to incorporate prior knowledge, thereby enhancing the agent's environmental understanding and achieving an effective balance between exploration and exploitation. \textbf{Leveraging the powerful language processing capabilities of LLMs, our framework removes the need for highly structured prior knowledge, thereby allowing broader utilization of human knowledge sources. This distinct advantage sets our approach apart from other methods.} Moreover, the text generated by LLMs often reflects structural patterns of the real world, embedding common-sense knowledge about various aspects of human reasoning and intuition \cite{rytting2021leveraging, gurnee2023language, HOU2025113624}.

\textbf{Compared to some recent methods \cite{yin2023lumos, yao2023retroformer, zhu2023ghost, deng2024promoting, deng2025cognidual, DU2025113546} that directly use LLMs as agents within the RL process, the advantage of our approach lies in the fact that LLMs are only required during the training phase to assist the agent in learning.} Once training is complete, our agent can be deployed independently without LLMs. In contrast to agents utilizing LLM kernels, conventional RL agents founded upon multilayer perceptrons or convolutional neural networks exhibit a comparative advantage regarding computational resource utilization \cite{xi2023rise, LI2025113471}. These advantages become particularly relevant in large-scale application scenarios and latency-sensitive environments.

Our interaction with LLMs involves LLMs processing environmental information and scoring agent behavior to guide exploration and exploitation through reward-shifting mechanisms. For certain non-text environments, we employ Visual Instruction Tuning \cite{liu2023llava} to align visual information with the projection matrices of LLMs. The rationale for this approach will be detailed in Section \ref{sec:housekeep}. Additionally, our method aligns with a key principle: reward shifting is equivalent to modifying the initialization of the Q-function, effectively balancing the exploration and exploitation aspects of RL \cite{sun2022exploit}.

We conducted extensive experiments across diverse settings and environments to evaluate our proposed method. The results demonstrate that our approach effectively leverages prior knowledge, significantly reducing the computational resources required for model training compared to baseline methods. Additionally, we assessed the performance of various LLMs within our framework, providing insights into their inferential capabilities in this context. To validate the versatility of our approach, we implemented our framework in Housekeep \cite{kant2022housekeep}, a simulated robotic environment, and evaluated its performance on complex robotic tasks. Furthermore, we applied our framework to Google's industrial-grade recommendation algorithm, SlateQ \cite{ie2019slateq}.

Our contributions can be summarized as follows:
\begin{itemize}
    \item We propose a novel framework for balancing exploration and exploitation in RL by leveraging LLMs. This framework addresses the exploration-exploitation dilemma and effectively guides the behavior of agents.
    \item We validate our proposed method across various RL environments and algorithms. Our approach significantly reduces the training cost of RL models while maintaining generality and ease of use.
    \item We demonstrate the effectiveness of our method in an industrial application context, providing a practical and straightforward solution to reduce the training cost of RL models for industry practitioners.
\end{itemize}

\section{Related Work}
Our research has significant connections to various preceding studies. Within the realm of reinforcement learning, there has been a comprehensive exploration of strategies to balance exploration and exploitation, a concept that resonates with our methodology. Our approach innovates within this field by harmonizing reinforcement learning principles with the capabilities of LLMs. This harmonization broadens our comprehension and application of balancing strategies, empowering us to tackle decision-making challenges in intricate settings more effectively. Moreover, our study draws inspiration from contemporary research on the evolution of LLMs, which has laid the theoretical groundwork and catalyzed noteworthy advancements in our inquiry.
\subsection{Exploration-Exploitation Trade-off in RL}

Numerous strategies addressing the RL exploration-exploitation dilemma have been proposed, encompassing various tasks. Established approaches such as the $\epsilon$-greedy \cite{sutton2018reinforcement}, Softmax \cite{sutton2018reinforcement}, Upper Confidence Bound (UCB) \cite{jin2018q}, and Thompson sampling \cite{russo2018tutorial} methods are widely recognized. Furthermore, entropy-based RL frameworks have seen progressive advancements \cite{haarnoja2018soft,ahmed2019understanding}, facilitating diverse exploration pathways in Markov Decision Processes (MD\-Ps), rendering them highly effective in continuous action spaces. The emergence of curiosity-driven RL strategies that leverage intrinsic curiosity as a reward signal \cite{pathak2017curiosity} has spurred extensive research \cite{burda2018large}. Notably, these approaches enable agents to interact with their environment and develop new skills. Despite their varying success in numerous contexts, these strategies typically exhibit static features, limiting their adaptability to dynamic settings. An innovative decoupling method was introduced, combining goal-specific and goal-agnostic exploration strategies for improved adaptability \cite{tarbouriech2021provably}. Additionally, deep covering options were investigated to enhance exploration efficiency without reliance on explicit state spaces or unwieldy action sets, courtesy of deep learning models and spectral analysis \cite{jinnai2019exploration}. Daochen \textit{et al.} \cite{zha2021rank} proposed a simplifying exploration method suitable for generative settings, fostering model-independent exploration through a ranking mechanism in environments with sparse rewards. While adept at accommodating changes autonomously, these approaches have restricted applicability due to their inattention to prior knowledge infusion, which could advance the equilibrium between exploration and exploitation. Our framework demonstrates versatility across diverse environments and RL algorithms, incorporating previously untapped prior knowledge to heighten agents' learning efficacy and minimize resource expenditure during training.

Hao \textit{et al.} \cite{sun2022exploit} theoretically prove that negative reward shifts aid exploration while positive reward shifts limit exploration. Ingmar \textit{et al.} \cite{schubert2021plan} introduce a novel reward shaping method that relaxes the strict optimality guarantee of potential-based reward shaping (PB-RS) while maintaining long-term behavioral consistency and enhancing sample efficiency compared to PB-RS. Inspired by these studies, we employed the reward shifts generated by LLMs to balance exploration and exploitation, enabling fine control of the impact on the existing RL process while making the overall learning process more efficient.

\subsection{Relevant Research in Large Language Models}

LLMs, particularly those based on the Transformer architecture \cite{vaswani2017attention}, such as OpenAI's GPT \cite{radford2018improving,openai2023gpt4}, Meta's Llama \cite{touvron2023llama}, Google's Bard \cite{thoppilan2022lamda}, LMSYS Org's Vicuna \cite{zheng2023judging}, and Anthropic's Claude \cite{bai2022training}, have widespread applications across various domains. Recent work \cite{qi-etal-2023-safer} in the field of LLMs has provided theoretical foundations for our research. \cite{rytting2021leveraging} demonstrates that the knowledge embedded within LLMs offers valuable inductive biases applicable to both traditional Natural Language Processing (NLP) tasks and non-traditional tasks involving training symbolic reasoning engines. The text generated by LLMs typically reflects structural patterns in the ``real world'', indicating that the weights within LLMs encode implicit common-sense knowledge related to various aspects of human reasoning and intuition. Through an analysis of learning representations in the Llama-2 series models across three spatial datasets and three temporal datasets, Wes \textit{et al.} \cite{gurnee2023language} presents evidence that LLMs form a coherent model of the data-generating process, which suggests that LLMs acquire structured knowledge regarding space and time fundamental dimensions. We posit that the weights embedded within LLMs inherently carry prior knowledge that can be utilized to guide agents' behavior.

\section{Methodology}
In this section, we present the overall structure and provide an in-depth exploration of the aspects relevant to prompts within our framework.

\subsection{Framework Structure}

RL methods are categorized into ``on-policy'' and ``off-policy'' based on how data is generated and processed. On-policy and off-policy methods are often viewed as distinct due to significant differences in their policy frameworks and algorithmic implementations in practice. These differences influence algorithm selection and optimization techniques. For instance, off-policy methods must address the importance of sampling issues associated with using data from non-target policies—a challenge not faced by on-policy methods. Broadly, however, on-policy methods can be seen as a subset of off-policy methods, where the behavior policy (which generates the data) aligns with the target policy (the policy under optimization). \textbf{Thus, off-policy definitions are inherently broader, encompassing all scenarios, even those where the learning and behavior policies coincide. All descriptions related to RL mentioned below refer to off-policy methods.} A common RL training process is as follows:

\begin{enumerate}
\item Initialization of the evaluation policy and the behavioral policy. The evaluation policy may be initialized as a stochastic policy, such as a random policy, while the behavioral policy may take the form of an $\epsilon$-greedy policy, incorporating a probability of random exploration.
\item The agent engages with the environment based on the behavioral policy, yielding training data in the form of state-action-reward-next state tuples. These data are then archived within an experience replay buffer.
\item Training data is sampled from the experience replay buffer, and the agent's parameters are updated based on the evaluation policy and the sampled data, employing techniques such as Temporal Difference (TD) learning or Monte Carlo methods.
\item Periodic evaluation of the evaluation policy's performance within the environment, with training termination contingent on the attainment of a predefined performance threshold.
\item The process iterates by returning to step 2, with periodic adjustments to the behavioral policy, such as the gradual reduction of $\epsilon$ in the case of an $\epsilon$-greedy policy.
\end{enumerate}

\textbf{To ensure the wide-ranging applicability of our improvements, we aim to preserve the fundamental principles of the original RL training process with minimal intervention.} LMGT introduces a specific modification to the second step, which involves adjusting the acquired experiences of the agent. We consider the LLM as the ``evaluator''. When the agent observes the environmental state, it selects an action based on the prevailing behavioral policy and communicates this action to the environment. We replicate and transmit both the observable state of the environment and the chosen action to the LLM. The LLM assesses the agent's actions and assigns a score, taking into account the prior knowledge that is embedded in its weights or introduced through the prompt (such as game rules). This score serves as a reward shift, which is incorporated into the reward generated by the environment itself. In contrast to the conventional RL process, LMGT involves the agent recording adjusted rewards instead of relying on the inherent rewards provided by the environment. The agent then learns from these adapted rewards to gain guidance from the LLMs.

In situations with sparse rewards, the agent faces challenges in accumulating information through trial and error. Hence, we employ the LLM to guide the agent, to avoid the pursuit of directions that have been previously determined as ``valueless'' based on prior knowledge, as indicated by a negative reward shift. The LLM assigns a positive reward shift for actions identified as ``valuable'' according to prior knowledge, encouraging the agent to focus on exploitation. While maintaining the traditional exploration-exploitation strategy from classical RL, the agent intensifies its exploration of actions neighboring those deemed ``valuable'' in the prior knowledge, increasing the likelihood of discovering the ``optimal'' action. Additionally, for actions not referenced in prior knowledge, the LLM assigns a ``0'' reward shift, allowing the agent to explore based on the original exploration policy.\par

The framework of LMGT is presented in Figure \ref{fig_lmgt}. For different tasks, the LLM provides various forms of reward shifts, guided by the principle that intricate tasks require more nuanced reward shifts, while simpler tasks require simpler reward shifts, using ``+1,'' ``0,'' and ``-1'' to represent ``approval'', ``neutral'', and ``disapproval'', respectively.\par

\begin{algorithm}[htbp]
\caption{Language Model Guided Trade-offs}
\SetKwData{State}{\texttt{state}}
\SetKwData{Action}{\texttt{action}}
\SetKwData{Reward}{\texttt{reward}}
\SetKwData{NextState}{\texttt{nextState}}
\SetKwData{Episode}{\texttt{episode}}
\SetKwData{Step}{\texttt{step}}
\SetKwData{Buffer}{\texttt{replayBuffer}}
\SetKwData{Policy}{\texttt{behaviorPolicy}}
\SetKwData{TargetPolicy}{\texttt{targetPolicy}}
\SetKwData{DiscountFactor}{\texttt{$\gamma$}}

\SetKwFunction{SelectAction}{\texttt{SelectAction}}
\SetKwFunction{TakeAction}{\texttt{TakeAction}}
\SetKwFunction{StoreTransition}{\texttt{StoreTransition}}
\SetKwFunction{SampleBatch}{\texttt{SampleBatch}}
\SetKwFunction{UpdateQNetwork}{\texttt{UpdateQNetwork}}

\SetKwInOut{Initialize}{Initialize}
\SetKwInOut{Input}{Input}
\SetKwInOut{Output}{Output}

\Initialize{
    Initialize the Reinforcement Learning environment\;
    Initialize the Q-value function or policy network\;
    Initialize the experience replay buffer\;
}

\Input{Total number of training episodes: $\text{N}$\;
       Maximum number of steps per episode: $\text{M}$\;
       Initial behavior policy: $\text{P}$\;
       Target policy: $\text{P' }$\;
}

\Output{Learned policy or value function}

\For{$\texttt{Episode}=1$ to $\texttt{N}$}{
    Reset the environment and obtain the initial state $s$\;
    \For{$\texttt{Step}=1$ to $\texttt{M}$}{
        Select an action $a$ using the behavior policy: $a = \SelectAction(s, P)$\;
        Execute action $a$ and observe the reward $r$ and the next state $s'$\;
        Incorporating $\left \langle s, a\right \rangle$ into LLM to obtain reward shift $\delta r$\;
        Let $r = r + \delta r$\;
        Store the transition $\left \langle s, a, r, s'\right \rangle$ in the experience replay buffer\;

        \If{Sufficient samples are available in the experience replay buffer}{
            Sample a batch of experiences from the replay buffer\;
            Compute the importance of sampling weights for the behavior policy\;
            Calculate the target values or estimates using the target policy\;
            Update the Q-value function or policy network using an Off-Policy learning algorithm\;
        }

        \If{$s$ is a terminal state}{
            Break from the current loop and move to the next episode\;
        }
    }
}
\label{alg1}
\end{algorithm}

\subsection{Prompt Design}
In this section, we will discuss the engineering methodology applied to optimize the performance of LLMs. The primary emphasis is placed on the performance attributes of LLMs, specifically concerning the magnitude of embedded prior knowledge in their weight configurations, as well as their capacity to comprehend and harness pre-existing knowledge about textual genres. The efficacy of the reward-shifting mechanism, generated by LLMs, fundamentally dictates the success of our approach and the extent of enhancement in comparison to the baseline.\par
Table \ref{tab:prompt_strategies} catalogs a detailed inventory of immediate enhancements utilized in our experimental design. It is important to note that the primary distinction between Zero-shot \cite{kojima2022large} and Baseline involves Zero-shot's integration of specific, task-related information into the prompt, which guides the LLM regarding the appropriate information to produce. The Name method could be perceived as perplexing. It involves attributing a name to an LLM in the prompt with the aim of improving performance. Nonetheless, our experimental results indicate that this technique does not yield any improvements. For additional details on the experiments, please see Section \ref{exp_1}. It is commonplace to deploy multiple prompt enhancements concurrently.\par
Furthermore, our prompt design is further categorized into two distinct classes: ``prior-knowledge-inclusive prompt statements'' and ``prior-knowledge-exclusive prompt statements''. The former class provides an all-encompassing evaluation of the LLMs' ability to harness their embedded prior knowledge, including their proficiency in leveraging prior knowledge presented in non-standard linguistic forms, such as natural language text. The latter class, on the other hand, exclusively investigates the LLMs' aptitude for exploiting implicit prior knowledge embedded within their model weights. 

Section \ref{exp} presents an elucidation of the effects of various prompt methods, along with a rationale for our methodological choices.

\section{Experiment} 
\label{exp}
The experiment is structured into three distinct parts. \textbf{In the initial phase, we scrutinize the benefits of our proposed framework over conventional approaches for addressing sparse reward challenges.} Specifically, we compare LMGT with Return Decomposition for Delayed Rewards (RUDDER) \cite{arjona2019rudder}, which is a novel RL approach for delayed rewards in finite MDPs. RUDDER's objective is to neutralize expected future rewards, thereby simplifying Q-value estimations to the average of immediate rewards. Despite RUDDER's expedited processing in scenarios with delayed rewards compared to traditional RL methods, it fails to incorporate prior knowledge—an area where LMGT particularly excels. Consequently, we anticipate LMGT to facilitate the expedited development of effective behavioral strategies by agents. \textbf{The second segment of the experiment evaluates our framework's versatility by applying it across diverse RL algorithms and environments to ascertain its efficacy.} Herein, we examined the enhancement in performance attributable to our framework across various RL algorithms, compared to some classic baselines. This phase further includes an assessment of the impact of different prompting techniques on our framework's performance and an exploratory evaluation of the reasoning capabilities of LLMs within our framework. \textbf{To ensure that the evaluation conclusions of our framework extend beyond synthetic settings, the final section investigates its application to and performance in complex robotic environments.} We conducted experiments using the Housekeep environment \cite{kant2022housekeep}. Housekeep is a simulated environment for embodied agents, where robots must perform housekeeping tasks by correctly placing out-of-place objects into appropriate containers.
\textbf{To ensure that the evaluation conclusions of our framework extend beyond synthetic settings, the final section investigates its practical applications and improvements.}  Specifically, we explore its integration with Google's SlateQ \cite{ie2019slateq}, a sophisticated recommendation algorithm that employs slate decomposition. This approach effectively manages the complexity of recommending multiple items simultaneously, addressing the challenge of large action spaces found in previous RL recommendation algorithms. This implementation was tested within a simulated environment on RecSim \cite{ie2019recsim}, a versatile platform for developing simulation environments for recommender systems (RSs), facilitating sequential user interactions.

\subsection{Experimental Settings}
For both LLM inference and agent training, we utilize a single NVIDIA A800-80G GPU. We adhere to the recommended settings by Llama for precise inference, which encompass a temperature of 0.7, top\_p of 0.1, a repetition penalty of 1.18, and top\_k of 40.

\subsection{Comparison Experiments with Traditional Exploration-Exploitation Trade-off Methods}

Reinforcement learning (RL) presents two fundamental challenges: the credit assignment problem for delayed rewards and the efficiency of exploration in complex environments. The former concerns the attribution of sparse, delayed rewards to preceding actions in a decision sequence, while the latter addresses the optimal balance between exploration and exploitation in unknown environments. Addressing these challenges has led researchers to develop various approaches, notably reward redistribution methods such as RUDDER and intrinsic motivation techniques such as NGU (Never Give Up). In this section, we empirically evaluate LMGT's efficacy in addressing these core challenges through comparative experiments with these representative algorithms.

\subsubsection{Comparison with RUDDER's Reward Redistribution Method}

RUDDER (Return Decomposition for Delayed Rewards) represents a specialized algorithm designed to address delayed reward problems. It employs a reward redistribution function that decomposes cumulative delayed rewards and attributes them to key preceding decision points. This approach is particularly effective for scenarios with challenging credit assignment, exemplified by the pocket watch repair task utilized in our experiments. In this task, agents must determine whether to repair watches of specific brands by conducting cost-benefit analyses between known selling prices and uncertain repair costs. The task embodies a classic delayed reward problem, as the profitability of repair decisions becomes apparent only after all associated costs are finalized.

\begin{table}[htbp]
\centering
\caption{\textbf{Performance comparison in watch repair task.} Evaluation of LMGT against baseline methods. All reported times include the inference time of the language model.}
\label{tab:watch_repair}
\begin{adjustbox}{width=0.35\textwidth}
\begin{tabular}{lrr}
\toprule
\textbf{Method} & \textbf{Episodes} & \textbf{Time (sec)} \\
\midrule
TD & 71,823 & 427 \\
MC & 221,770 & 530 \\
RUDDER & 2,029 & 171 \\
\textbf{LMGT+TD (ours)} & \textbf{417} & \textbf{114} \\
\bottomrule
\multicolumn{3}{l}{\footnotesize Lower values indicate better performance in both metrics.}\\
\end{tabular}
\end{adjustbox}
\end{table}

RUDDER was selected as a baseline for comparison because it represents a state-of-the-art method for addressing delayed reward problems and closely aligns with our research objectives. To ensure methodological consistency, LMGT was implemented with temporal difference (TD) learning, matching RUDDER's foundational approach. Experiments were conducted using multiple runs with random seeds {42, 43, 44, 45, 46}, and results were averaged. Performance was evaluated using two metrics: the number of training episodes required to achieve a profitable decision rate exceeding 90\% (reported as ``Episode'') and the total computational time needed for training (reported as ``Time''), inclusive of LLM inference time.

\begin{table*}[htbp]
\centering
\caption{\textbf{Performance of LMGT and baseline algorithms in hard-exploration Atari games.} Game scores at different training stages (measured in frames $\times 10^{10}$).}
\label{tab:performance_comparison}
\begin{adjustbox}{width=0.8\textwidth}
\begin{tabular}{@{}llcccc|cccc@{}}
\toprule
\multirow{2}{*}{Environment} & \multirow{2}{*}{Algorithm} & \multicolumn{4}{c|}{Early Training} & \multicolumn{4}{c}{Late Training} \\
\cmidrule(lr){3-6} \cmidrule(l){7-10}
 &  & 0 & 0.5 & 1.0 & 1.5 & 2.0 & 2.5 & 3.0 & 3.5 \\
\midrule
\multirow{3}{*}{Pitfall} 
 & R2D2 & 0.0 & 103.5 & 95.3 & 325.6 & 2786.3 & 3586.7 & 3548.6 & 3613.5 \\
 & NGU+R2D2 & 0.0 & 786.9 & 3233.3 & 4020.2 & 6400.0 & 5494.6 & 5120.0 & 5973.4 \\
 & \textbf{LMGT+R2D2 (ours)} & \textbf{0.0} & \textbf{1535.5} & \textbf{5486.3} & \textbf{6348.2} & \textbf{6125.6} & \textbf{6354.3} & \textbf{6535.2} & \textbf{6503.5} \\
\midrule
\multirow{3}{*}{\begin{tabular}[c]{@{}l@{}}Montezuma's\\Revenge\end{tabular}} 
 & R2D2 & 0.0 & 1235.6 & 2567.2 & 2465.8 & 2500.7 & 2754.6 & 2703.5 & 2687.2 \\
 & NGU+R2D2 & 0.0 & 4818.5 & 6369.2 & 7165.7 & 8236.5 & 9810.0 & 9966.7 & 9049.4 \\
 & \textbf{LMGT+R2D2 (ours)} & \textbf{0.0} & \textbf{5635.4} & \textbf{8957.3} & \textbf{11257.3} & \textbf{10985.7} & \textbf{12093.9} & \textbf{12463.6} & \textbf{12365.5} \\
\bottomrule
\multicolumn{10}{@{}l@{}}{\footnotesize LMGT is our proposed method. Best results are highlighted in bold.}\\
\end{tabular}
\end{adjustbox}
\end{table*}

As illustrated in Table \ref{tab:watch_repair}, experimental results demonstrate that LMGT substantially outperforms RUDDER on both metrics: LMGT requires only 417 training episodes and 114 seconds to develop a qualified strategy, whereas RUDDER necessitates 2029 episodes and 171 seconds. Notably, LMGT's proportional advantage in training episodes (approximately 79.4\% reduction) significantly exceeds its advantage in computational time (approximately 33.3\% reduction). This disparity suggests that the prior knowledge embedded in the language model effectively accelerates the value learning process, resulting in more efficient credit assignment.

\subsubsection{Comparison with NGU's Exploration-Driven Method}

With respect to the second critical challenge in reinforcement learning—exploration efficiency—NGU (Never Give Up) represents an advanced exploration methodology based on intrinsic motivation principles. NGU integrates novelty-driven intrinsic rewards with environmental extrinsic rewards, thereby encouraging continuous exploration of unknown state spaces while maintaining focus on high-value regions. While this approach demonstrates excellent performance in environments with sparse rewards, it nevertheless requires extensive environmental interactions to develop effective exploration strategies.

To assess LMGT's exploration efficiency advantages, we selected two Atari games characterized by highly sparse rewards: Pitfall and Montezuma's Revenge. These environments have been established as particularly challenging benchmarks for exploration algorithms due to their complex state spaces and extremely sparse reward mechanisms. For these experiments, R2D2 (Recurrent Replay Distributed DQN) served as the base algorithm, combined separately with NGU and LMGT. Performance was compared across various training frame quantities.

\begin{figure*}[htbp]
    \centering
    \subfloat[Pitfall]{\includegraphics[width=0.48\textwidth]{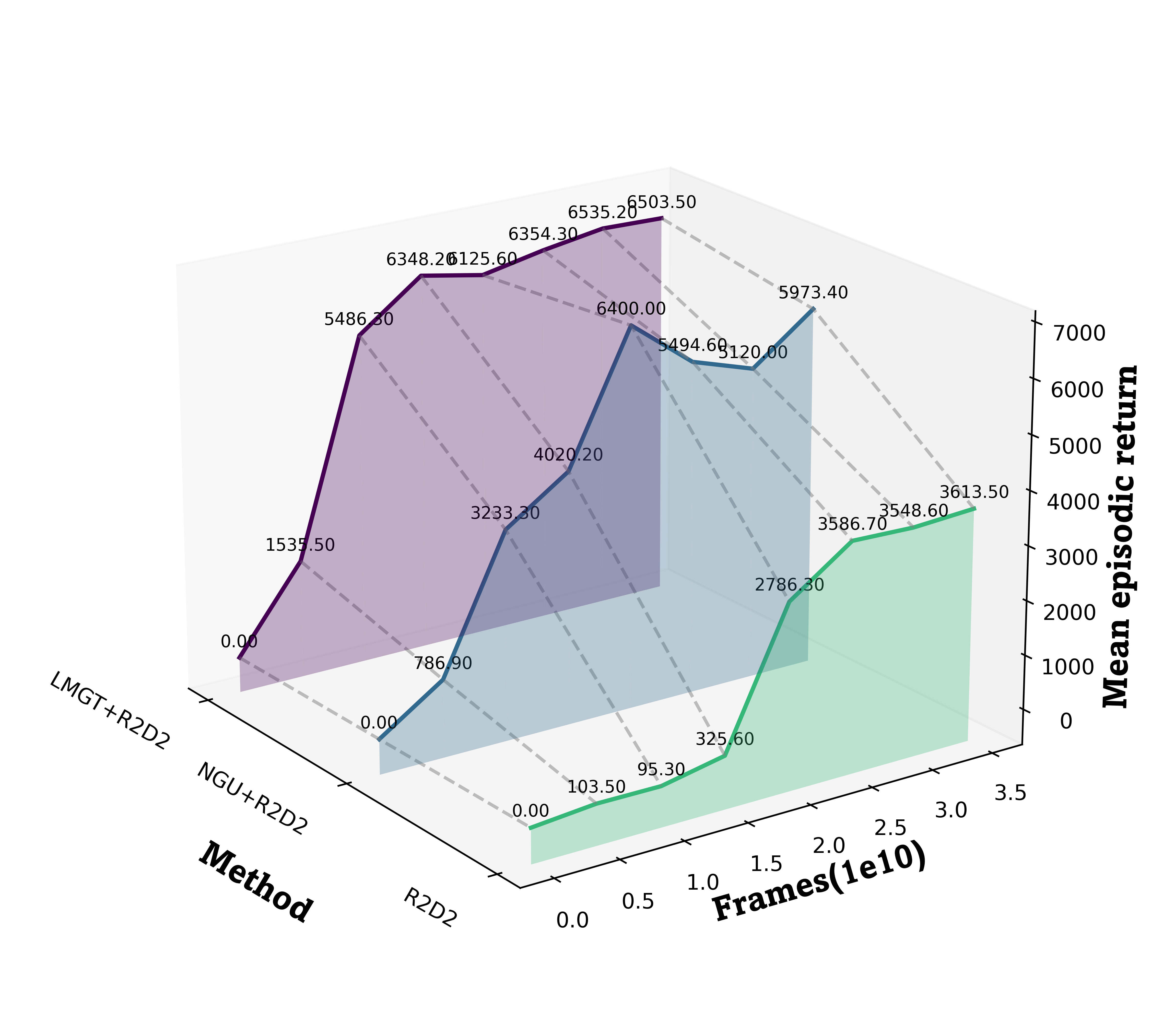}}\hfill
    \subfloat[Montezuma's Revenge]{\includegraphics[width=0.48\textwidth]{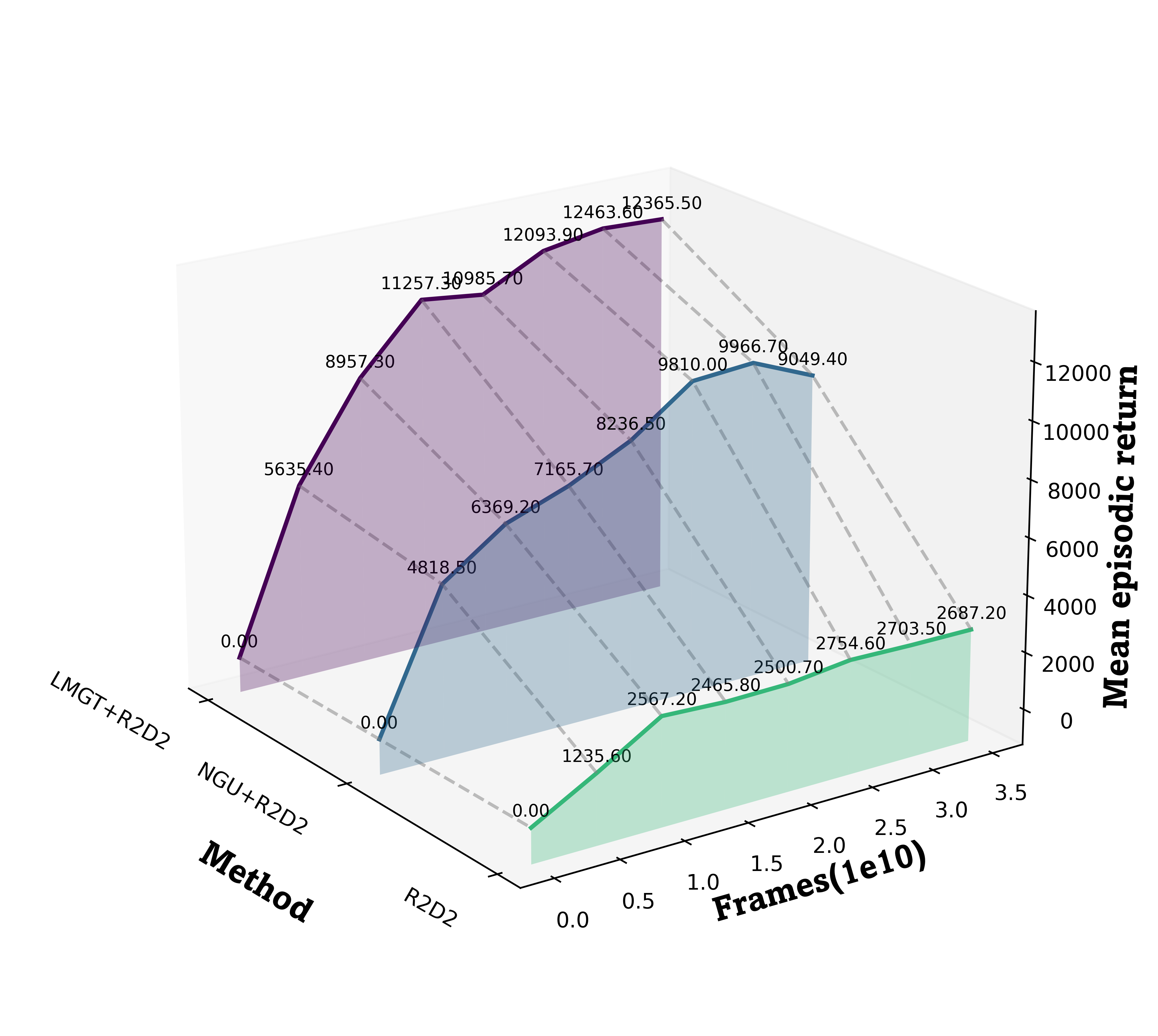}}
    
    \caption{\textbf{Performance comparison on challenging exploration Atari environments.} The figures present reward curves demonstrating our LMGT method's effectiveness in (a) Pitfall and (b) Montezuma's Revenge - two environments characterized by extremely sparse rewards and complex exploration requirements. Our approach shows significant performance gains over baseline methods in these notoriously difficult benchmarks where conventional RL algorithms typically struggle to make progress.}
    
    \label{fig:hard_exploration_games}
\end{figure*}

Experimental results (as shown in Figure \ref{fig:hard_exploration_games} and Table \ref{tab:performance_comparison}) reveal that LMGT+R2D2 progressively achieves superior performance in both environments as training progresses. After 3.5×10\^10 frames of training, LMGT+R2D2 attains an average reward of 6503.5 in the Pitfall environment, surpassing NGU+R2D2 (5973.4) by approximately 8.9\% and exceeding the baseline R2D2 implementation (3613.5) by approximately 80.0\%. In the more challenging Montezuma's Revenge environment, the performance differential increases markedly: LMGT+R2D2 achieves a reward of 12365.5, exceeding NGU+R2D2 (9049.4) by approximately 36.6\% and outperforming baseline R2D2 (2687.2) by a factor of 4.6.

Perhaps more significant are the performance disparities observed during early training phases. With merely 0.5×10\^10 training frames, LMGT+R2D2 achieves a reward of 1535.5 in Pitfall, nearly double that of NGU+R2D2 (786.9). Similarly, in Montezuma's Revenge, LMGT+R2D2 (5635.4) substantially outperforms NGU+R2D2 (4818.5). These early-stage advantages indicate that LMGT, leveraging prior knowledge embedded in language models, can more rapidly identify high-value regions, thereby significantly enhancing exploration efficiency.

\subsubsection{Comprehensive Analysis and Discussion}

Our comparative experiments with RUDDER and NGU demonstrate LMGT's substantial advantages in addressing both core reinforcement learning challenges. These advantages derive from LMGT's distinctive operational mechanism: language models not only provide structural environmental knowledge but also guide exploration through intelligent reward shifting, simultaneously optimizing both credit assignment processes and exploration strategies.

For delayed reward problems, LMGT utilizes language model-based action value assessment to provide more direct credit assignment mechanisms, enabling agents to rapidly identify critical decision points and substantially reducing required training episodes. For exploration challenges, LMGT employs language model evaluations of state-action pairs to assign positive reward shifts to potentially valuable trajectories, thus avoiding inefficient exploration and achieving more optimal exploration-exploitation balance in complex environments.

It is particularly noteworthy that LMGT's advantages become increasingly pronounced as task complexity increases. This scalability arises because language models based on Transformer architectures maintain relatively constant inference speeds regardless of task complexity, whereas traditional methods typically require exponentially increasing exploration time as environments become more complex. This characteristic renders LMGT especially suitable for high-dimensional environments and long-horizon decision problems.

In conclusion, by integrating the cognitive capabilities of large language models with reinforcement learning frameworks, LMGT provides an innovative approach that simultaneously addresses credit assignment and exploration efficiency challenges. This integration opens new avenues for reinforcement learning applications in resource-constrained scenarios, potentially transforming how complex sequential decision problems are approached.

\subsection{Evaluation of LMGT among Various Reinforcement Learning Algorithms and Environments}
\label{exp_1}
\subsubsection{Main experiment}
\label{main exp}
\begin{figure*}[htbp]  
    \centering
    \subfloat[Cart Pole(DQN)]{\includegraphics[width=0.25\textwidth]{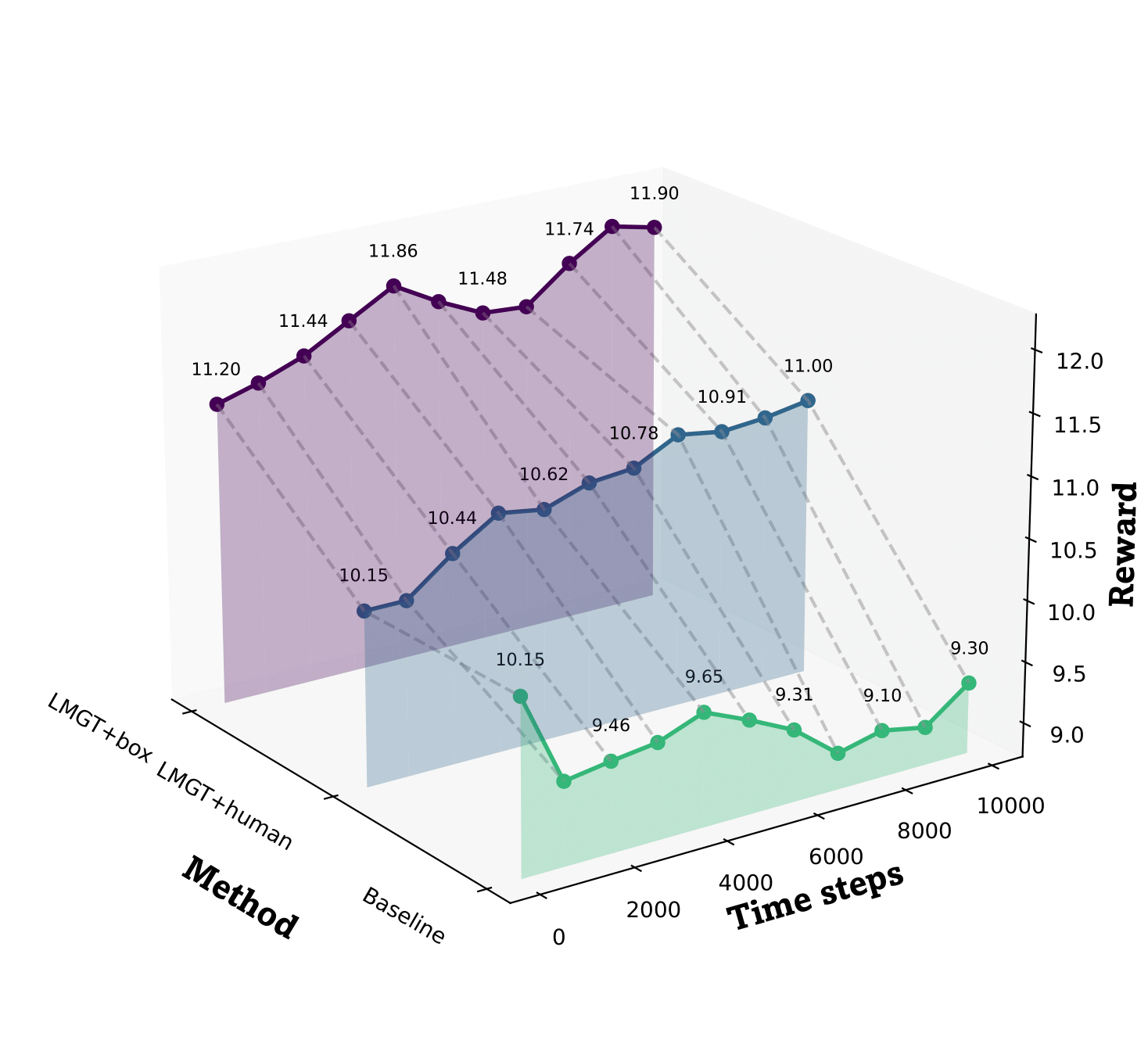}}\hfill
    \subfloat[Cart Pole(PPO)]{\includegraphics[width=0.25\textwidth]{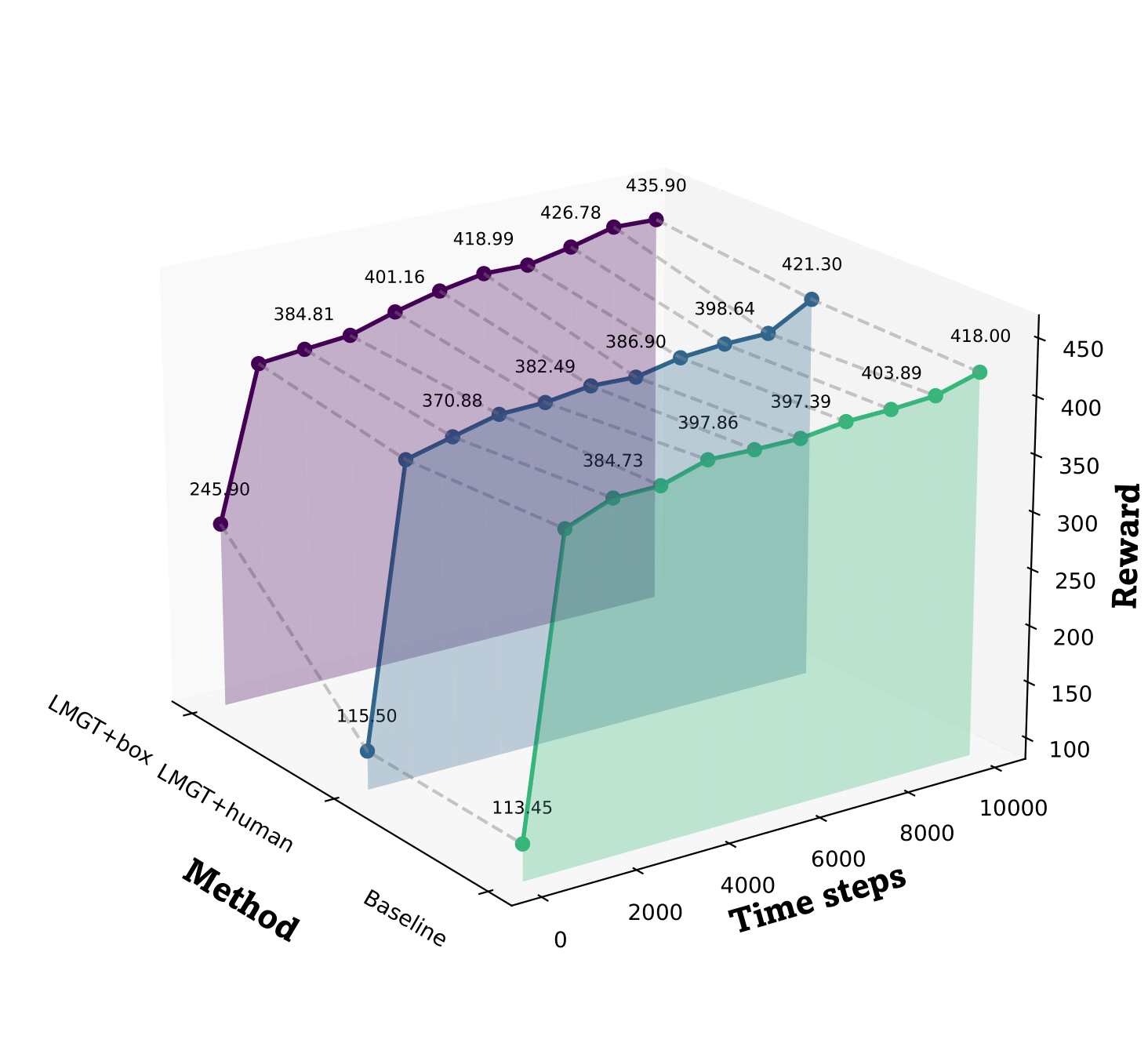}}\hfill
    \subfloat[Cart Pole(A2C)]{\includegraphics[width=0.25\textwidth]{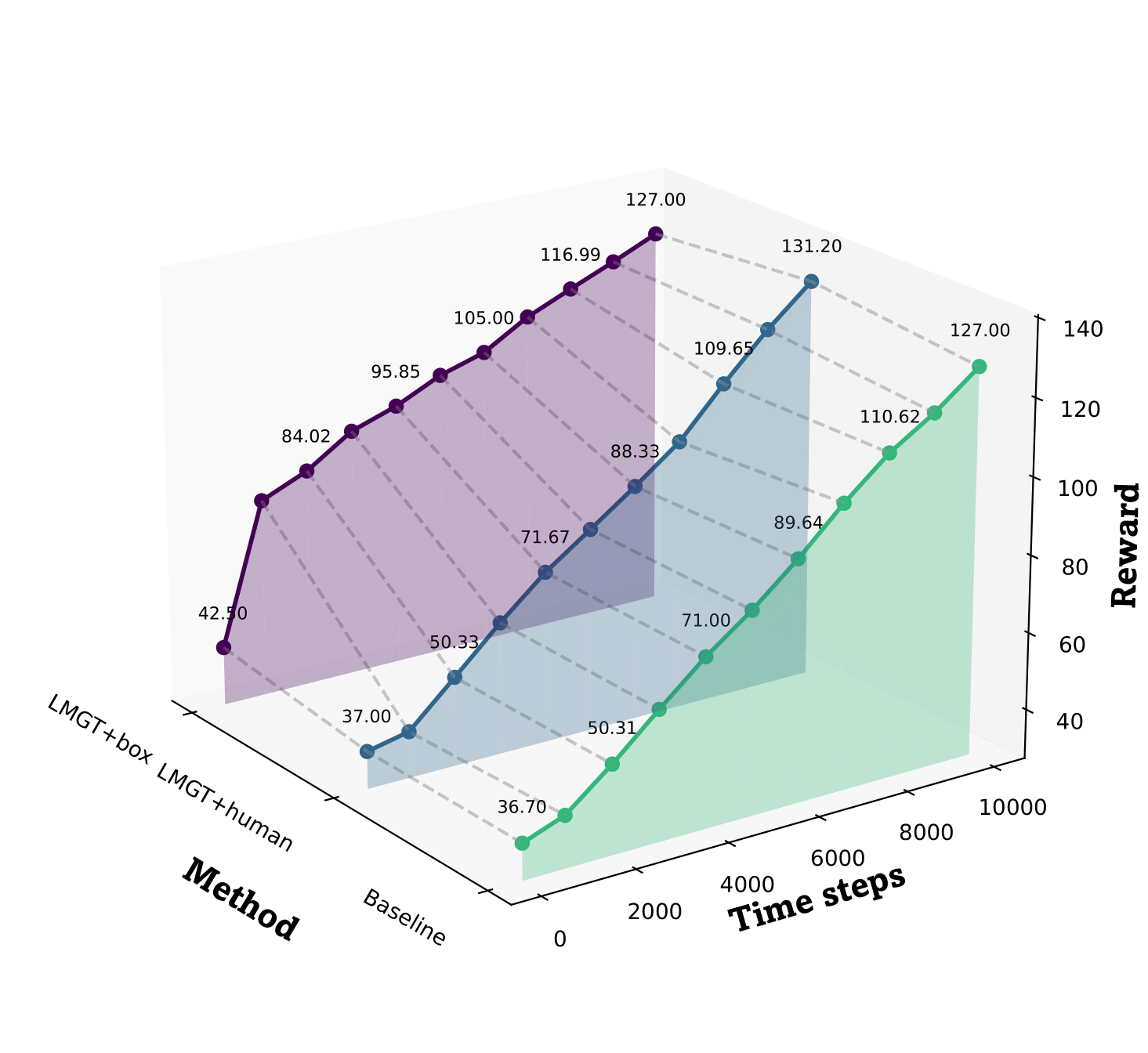}}
    \vspace{-12pt} 
    \subfloat[Pendulum(SAC)]{\includegraphics[width=0.25\textwidth]{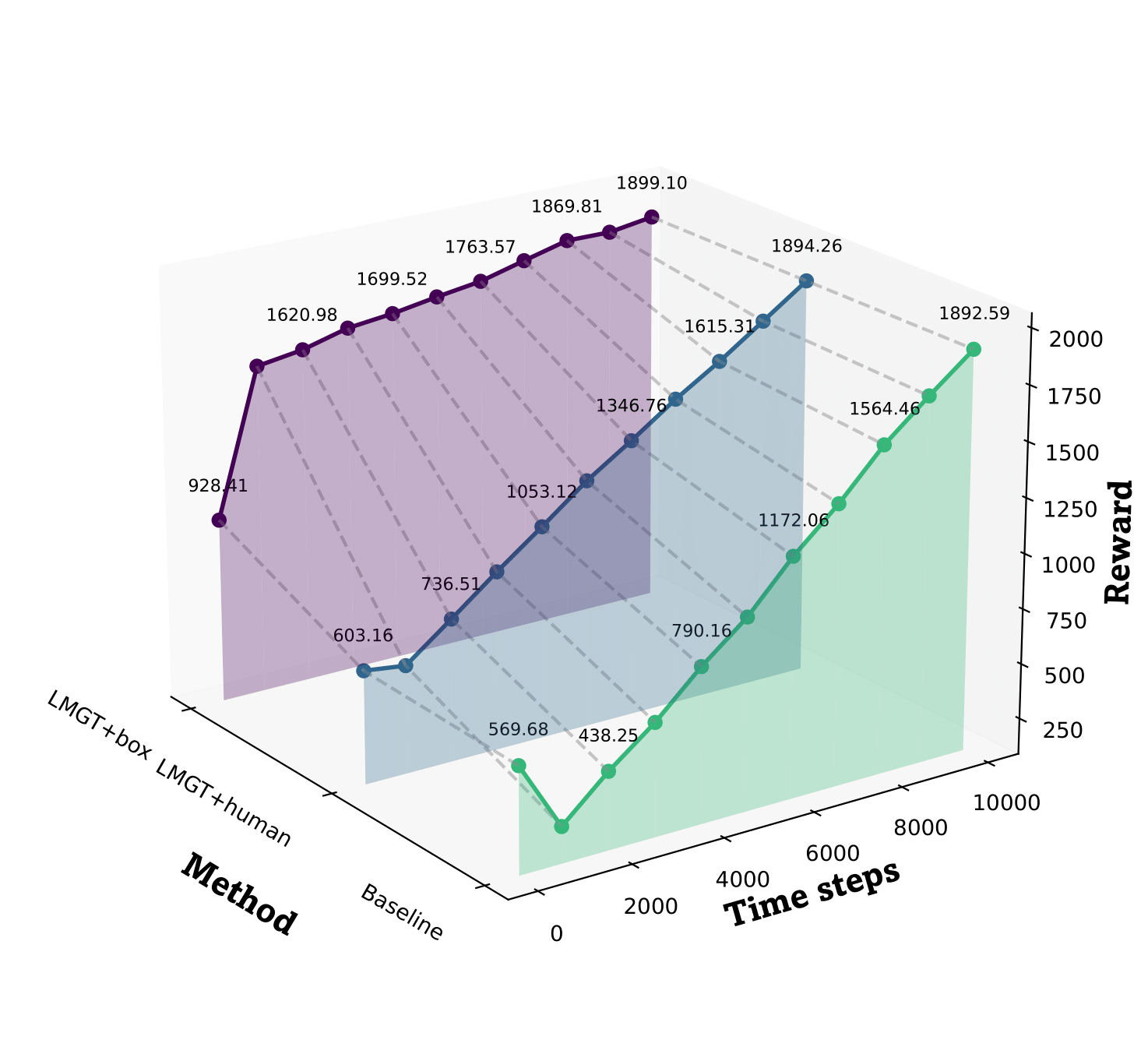}}\hfill
    \subfloat[Pendulum(TD3)]{\includegraphics[width=0.25\textwidth]{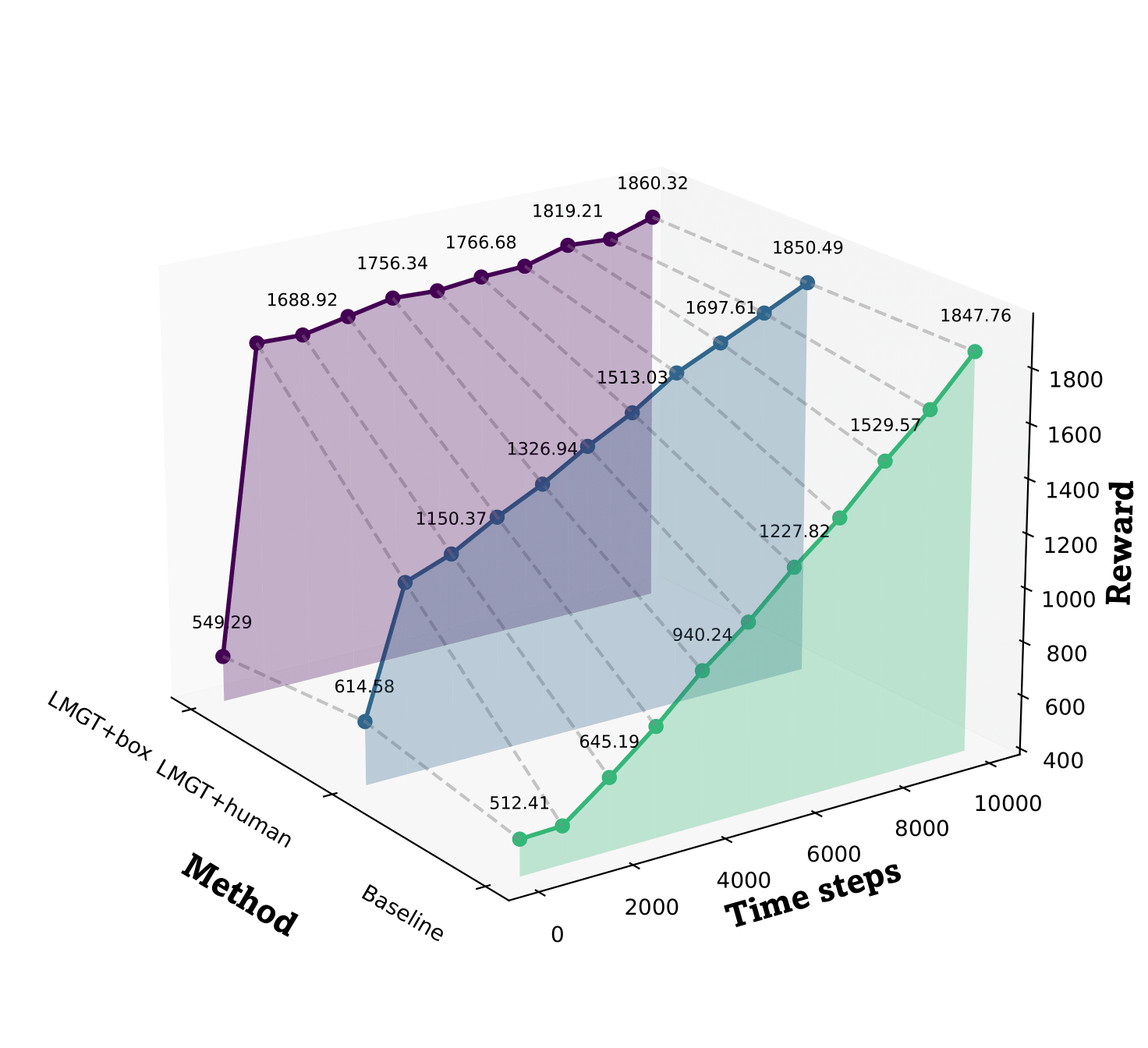}}\hfill
    \subfloat[Pendulum(PPO)]{\includegraphics[width=0.25\textwidth]{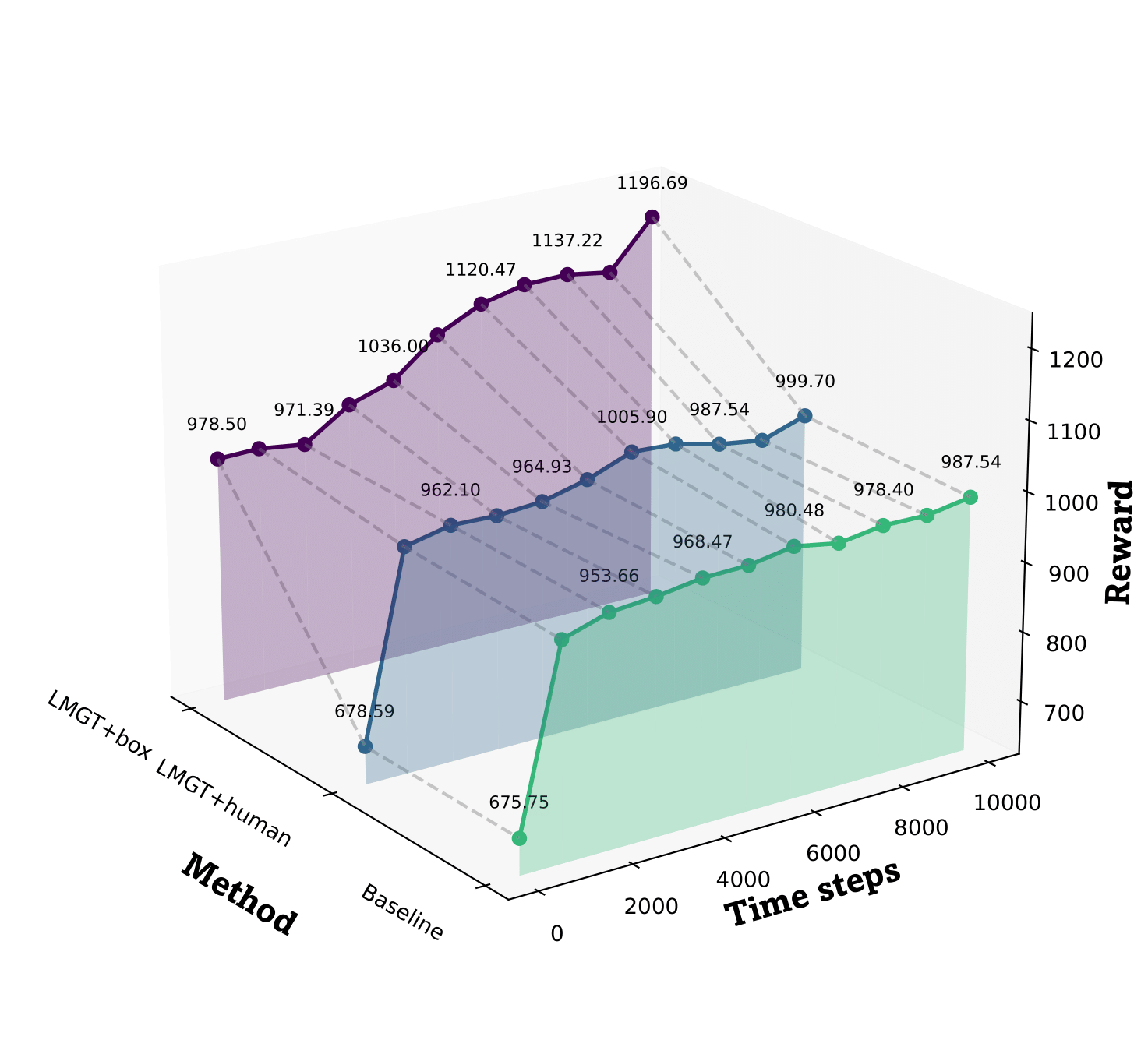}}\hfill
    \subfloat[Pendulum(A2C)]{\includegraphics[width=0.25\textwidth]{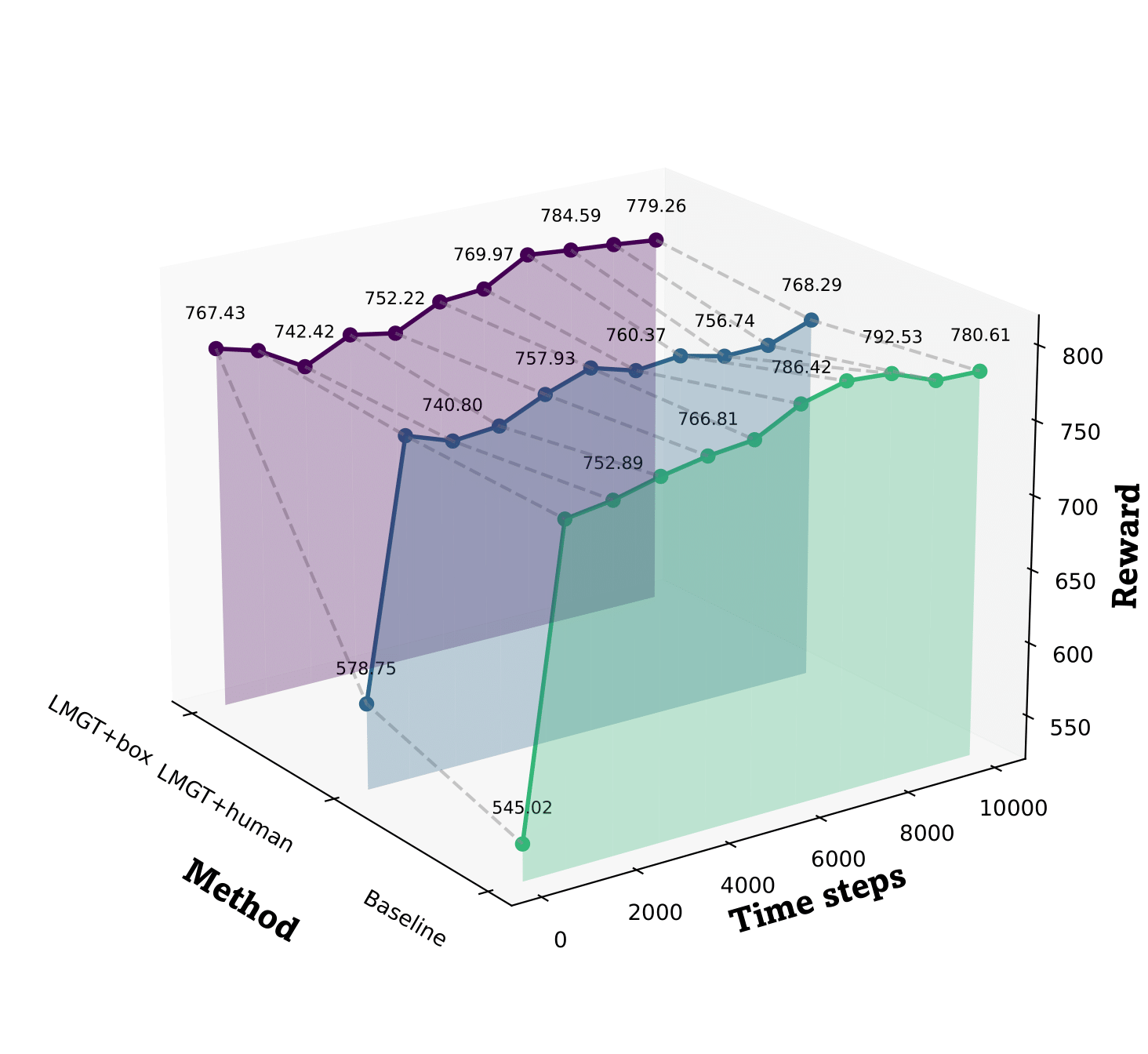}}
    \caption{\textbf{Results of experiments conducted across varying settings.} It is important to note that all rewards in the Pendulum environment are negative. To enhance the visualization, each reward value shown in the graphs has been increased by an offset of 2000.}  
    \label{fig:mainresults}
\end{figure*}

This section undertakes a comprehensive evaluation of the efficacy of our LMGT framework across various RL environments, employing diverse RL algorithms. Table \ref{tab:boosted_rewards} presents a comprehensive overview of our experimental results. \textbf{The values in the table indicate the performance gain in terms of reward achieved by LMGT compared to the baseline method, referred to as the `boosted reward'.} Notably, red figures denote scenarios where our proposed method underperforms compared to the baseline.  \textbf{Please be advised that all rewards presented in the results correspond to the environments' intrinsic rewards.} The environments are identified with their observable states furnished to the LLMs in two distinct formats: a standardized numerical representation, denoted as ``box'' (e.g., a tuple encapsulating information on object positions), and a more intuitively comprehensible visual format referred to as ``human'' (such as a screenshot of the current frame).  Our metric for assessing our approach against baseline methods is the ``average reward of the model after a fixed number of training time steps''. Specifically, agents are trained separately using our method and baseline techniques within the same environment, and the trained weights are preserved after a predefined number of time steps. Subsequently, we evaluate the performance of models trained using different methods, employing an equivalent number of training time steps in the same environment, while comparing their average rewards. For a more intuitive representation of the results, Figure \ref{fig:mainresults} illustrates the learning trajectories of agents under various experimental conditions, providing a visual comparison of performance over time steps. \par

Throughout these experiments, we maintained a consistent choice of LLM and prompt techniques. Specifically, two prompt methods were employed: CoT and Zero-shot prompt, to formulate our prompts. The 4-bit quantized version of the Vicuna-30B model \cite{vicuna2023}, with GPTQ quantization \cite{frantar2022gptq}, was utilized as our guiding LLM within our framework. This model is utilized to assess the quality of agent behavior in distinct environmental states. We contend that this configuration optimizes the performance of our framework, and we will delve into the influence of different prompt techniques and LLMs on the framework's performance in other parts of this section.\par

RL environments are seldom conveyed through purely textual descriptions; thus, LLMs necessitate multimodal capabilities to process such information. Common LLMs such as Llama, Llama2, and Vicuna do not inherently support multimodal functionality. \textbf{To address this limitation, we adopted a pipeline model approach, where multiple single-modal models work synergistically, with each model responsible for processing specific data types and passing results to the next model to accomplish tasks.} In our experiments, we integrated LLaVA \cite{liu2023llava} as the image processing model preceding the LLM. Therefore, in the aforementioned experiments, LLaVA was integrated with the Vicuna-30B model and operated collaboratively, equipping our ``scorer'' with image processing capabilities. Notably, our integration method does not simply convert the environment into text through an image captioning model before transmission to LLMs. Instead, it directly inputs embeddings representing the environmental state into the LLMs. This approach allows us to preserve more meaningful information to support LLMs in their decision-making processes. \par

Table \ref{tab:boosted_rewards} illustrates that our framework consistently outperforms baseline methods across a majority of environments and various RL algorithms. It effectively achieves a trade-off between exploration and exploitation in RL methods, enabling agents to acquire skills more rapidly, thus leading to cost savings during training. Moreover, we observed that our framework's performance is relatively inferior in tasks necessitating the utilization of pipeline models to process visual information compared to tasks that exclusively involve text information processing. In essence, if Vicuna-30B is required to handle additional image information from LLaVA, its performance tends to deteriorate. An intriguing observation proposed in \cite{clavie2023large} suggests that attempting to enforce strict adherence of the LLM to response templates results in reduced performance across all scenarios. We hypothesize that both these scenarios signify a degradation in LLM performance in multitask settings \cite{hendrycks2020measuring}. Within our framework, ``understanding extracted image information'' and ``assigning scores to agent behavior based on a combination of different information'' represent distinct tasks, while the phenomenon mentioned in \cite{clavie2023large} pertains to ``providing responses based on prompts'' and ``formatting responses as required'' as two separate tasks.\par

\begin{table*}[htbp]
\centering
\caption{\textbf{Performance gains achieved by LMGT across different settings.} Values represent the boosted reward compared to baseline methods. Negative values (shown in \textcolor{red}{red}) indicate performance degradation.}
\label{tab:boosted_rewards}
\begin{adjustbox}{width=0.65\textwidth}
\begin{tabular}{llrrrrrr}
\toprule
\multirow{3}{*}{\textbf{Environment}} & \multirow{3}{*}{\textbf{Algorithm}} & \multicolumn{6}{c}{\textbf{Observable Environmental State Format}} \\
\cmidrule{3-8}
& & \multicolumn{3}{c}{Box} & \multicolumn{3}{c}{Human} \\
\cmidrule(lr){3-5} \cmidrule(l){6-8}
& & \multicolumn{1}{c}{n=100} & \multicolumn{1}{c}{n=1000} & \multicolumn{1}{c}{n=10000} & \multicolumn{1}{c}{n=100} & \multicolumn{1}{c}{n=1000} & \multicolumn{1}{c}{n=10000} \\
\midrule
\multirow{3}{*}{Cart Pole} 
& DQN & 1.05 & 1.90 & 2.60 & 0.00 & 0.75 & 1.70 \\
& PPO & 132.45 & 11.85 & 17.90 & 2.05 & \textcolor{red}{-8.50} & 3.30 \\
& A2C & 5.80 & 38.20 & 0.00 & 0.30 & \textcolor{red}{-1.40} & 4.20 \\
\midrule
\multirow{4}{*}{Pendulum} 
& SAC & 358.73 & 1338.20 & 6.51 & 33.48 & 326.24 & 1.67 \\
& TD3 & 36.88 & 1177.35 & 12.56 & 102.17 & 569.88 & 2.73 \\
& PPO & 302.75 & 47.03 & 209.15 & 2.84 & 14.68 & 12.16 \\
& A2C & 222.41 & 11.86 & \textcolor{red}{-1.35} & 33.73 & 3.12 & \textcolor{red}{-12.32} \\
\bottomrule
\multicolumn{8}{l}{\footnotesize Higher values indicate greater improvement by LMGT over the baseline method.}\\
\end{tabular}
\end{adjustbox}
\end{table*}

We also investigated the influence of different prompt methods on the performance of our framework. Similar to the previous experiments, while keeping other variables constant, we continued to employ the 4-bit quantized version of the Vicuna-30B model as our LLM and the A2C algorithm as our RL technique. We conducted tests on two representative environments, and the experimental results are presented in Table \ref{tab:prompt_strategies}. It is assumed that prompts in the table all inherently contain prior knowledge. For both simple (Cart Pole) and complex (Blackjack) environments, the most effective prompt method was found to be CoT. CoT particularly excelled in enhancing performance for complex tasks. We discovered that the model often overlooked the provided information and resulted in a uniform outcome unless explicitly instructed to employ hierarchical thinking in challenging tasks. Furthermore, we observed that merely assigning a simple name to the model scarcely enhanced its performance.\par

An intriguing observation emerged when comparing prompt methods on our task: the ``Zero-shot prompt'' method outperformed the ``Few-shot prompt'' method. Few-shot prompts often led the Vicuna-30B model to generate results with a sense of ``illusion''. Vicuna-30B frequently produced arbitrary extensions based on the provided examples. Furthermore, we observed that incorporating prior knowledge into the prompts can lead to an improvement in the performance of our framework, despite the fact that the weights within the Vicuna-30B model already encompass the requisite prior knowledge for addressing the challenges presented by the environment.\par

\begin{table*}[htbp]
\centering
\caption{\textbf{Impact of various prompt strategies on reinforcement learning performance.} Higher values indicate better performance, with the best results in each column shown in bold.}
\label{tab:prompt_strategies}
\begin{adjustbox}{width=0.7\textwidth}
\begin{tabular}{llrrr}
\toprule
\textbf{Environment} & \textbf{Prompt Strategy} & \multicolumn{3}{c}{\textbf{Time Steps}} \\
\cmidrule(lr){3-5}
& & n=100 & n=1000 & n=10000 \\
\midrule
\multirow{7}{*}{Cart Pole} 
& Baseline & 37.70 & 42.70 & 125.90 \\
& CoT \cite{wei2022chain} & 42.10 & 74.00 & 126.00 \\
& Zero-shot prompt \cite{kojima2022large} & 38.70 & 68.90 & 126.00 \\
& Few-shot prompt \cite{touvron2023llama} & 38.10 & 65.00 & 125.10 \\
& Name \cite{clavie2023large} & 37.10 & 42.90 & 125.90 \\
& CoT+Zero-shot prompt (w/o prior knowledge) & 42.00 & 77.10 & 126.10 \\
& CoT+Zero-shot prompt (w/ prior knowledge) & \textbf{42.50} & \textbf{78.90} & \textbf{127.00} \\
\midrule
\multirow{7}{*}{Blackjack} 
& Baseline & -0.20 & 0.20 & 0.32 \\
& CoT \cite{wei2022chain} & 0.10 & 0.28 & 0.45 \\
& Zero-shot prompt \cite{kojima2022large} & 0.10 & 0.28 & 0.32 \\
& Few-shot prompt \cite{touvron2023llama} & -0.20 & 0.20 & 0.32 \\
& Name \cite{clavie2023large} & -0.20 & 0.20 & 0.33 \\
& CoT+Zero-shot prompt (w/o prior knowledge) & 0.00 & 0.25 & 0.40 \\
& CoT+Zero-shot prompt (w/ prior knowledge) & \textbf{0.12} & \textbf{0.30} & \textbf{0.45} \\
\bottomrule
\multicolumn{5}{p{0.9\textwidth}}{\footnotesize CoT = Chain of Thought. ``w/ prior knowledge'' indicates the inclusion of domain-specific knowledge in the prompt, while ``w/o'' indicates its exclusion.}\\
\end{tabular}
\end{adjustbox}
\end{table*}

\begin{figure*}[htbp]  

    \centering
    \subfloat[Task1]{\includegraphics[width=0.25\textwidth]{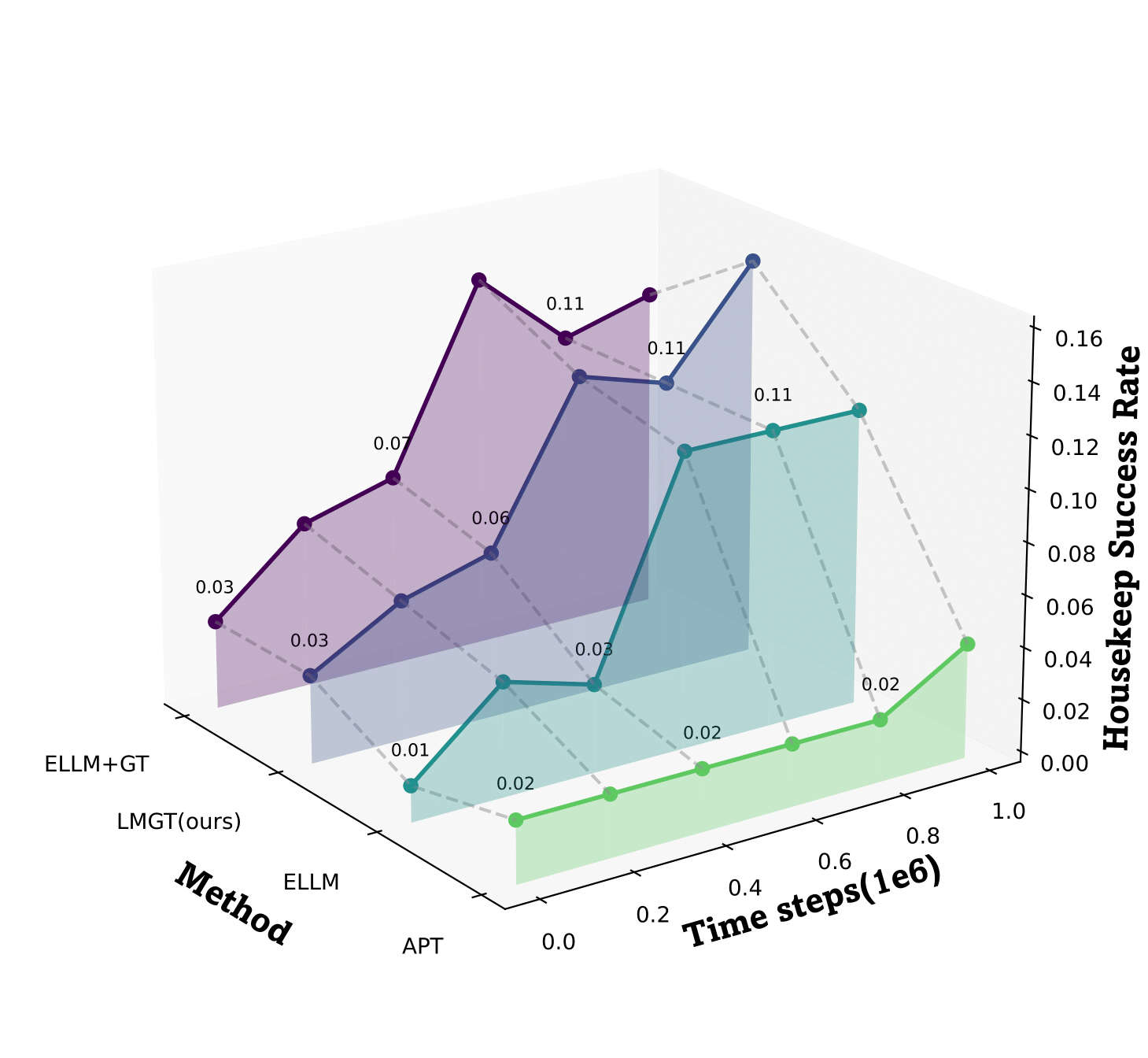}}\hfill
    \subfloat[Task2]{\includegraphics[width=0.25\textwidth]{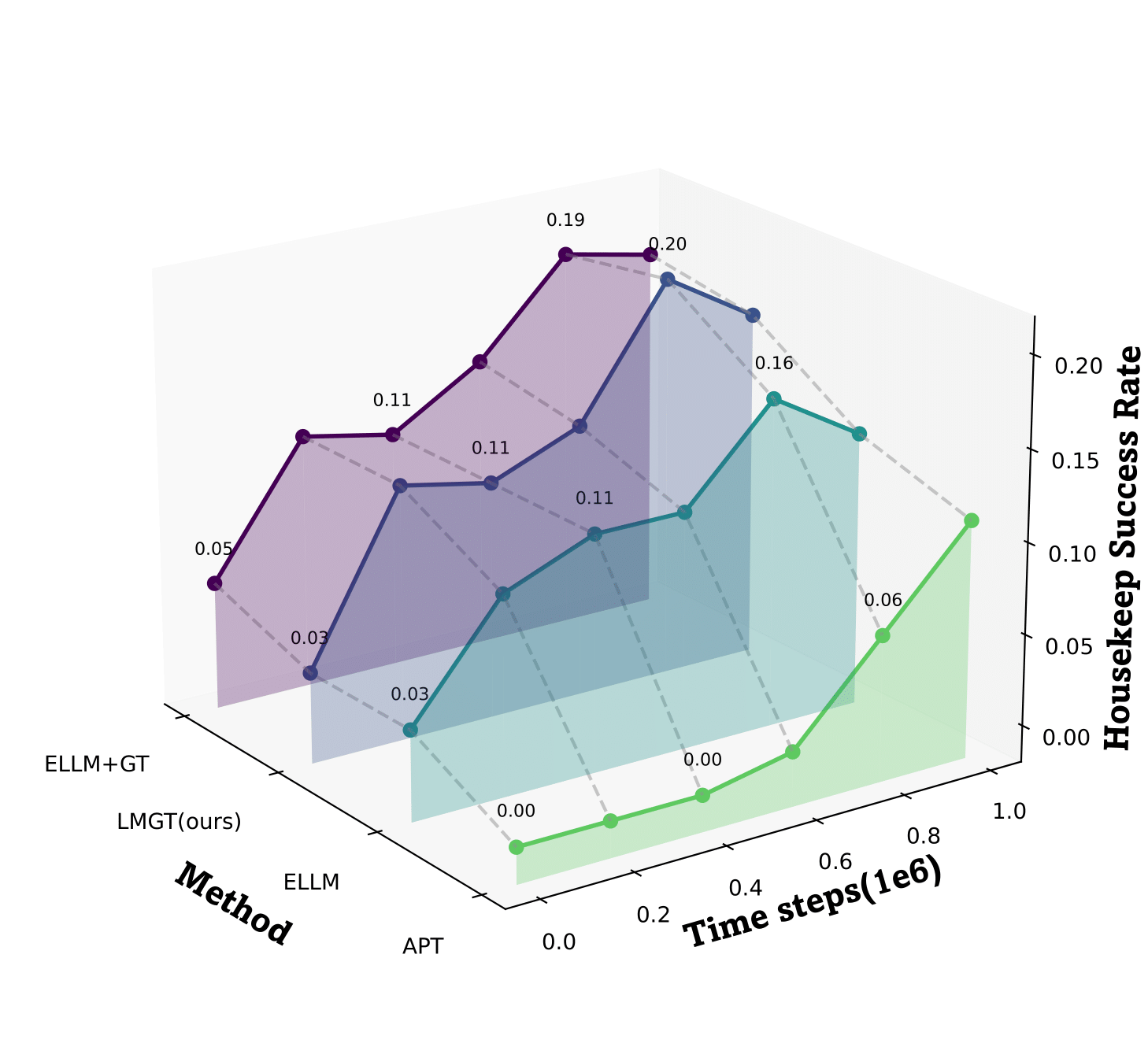}}\hfill
    \subfloat[Task3]{\includegraphics[width=0.25\textwidth]{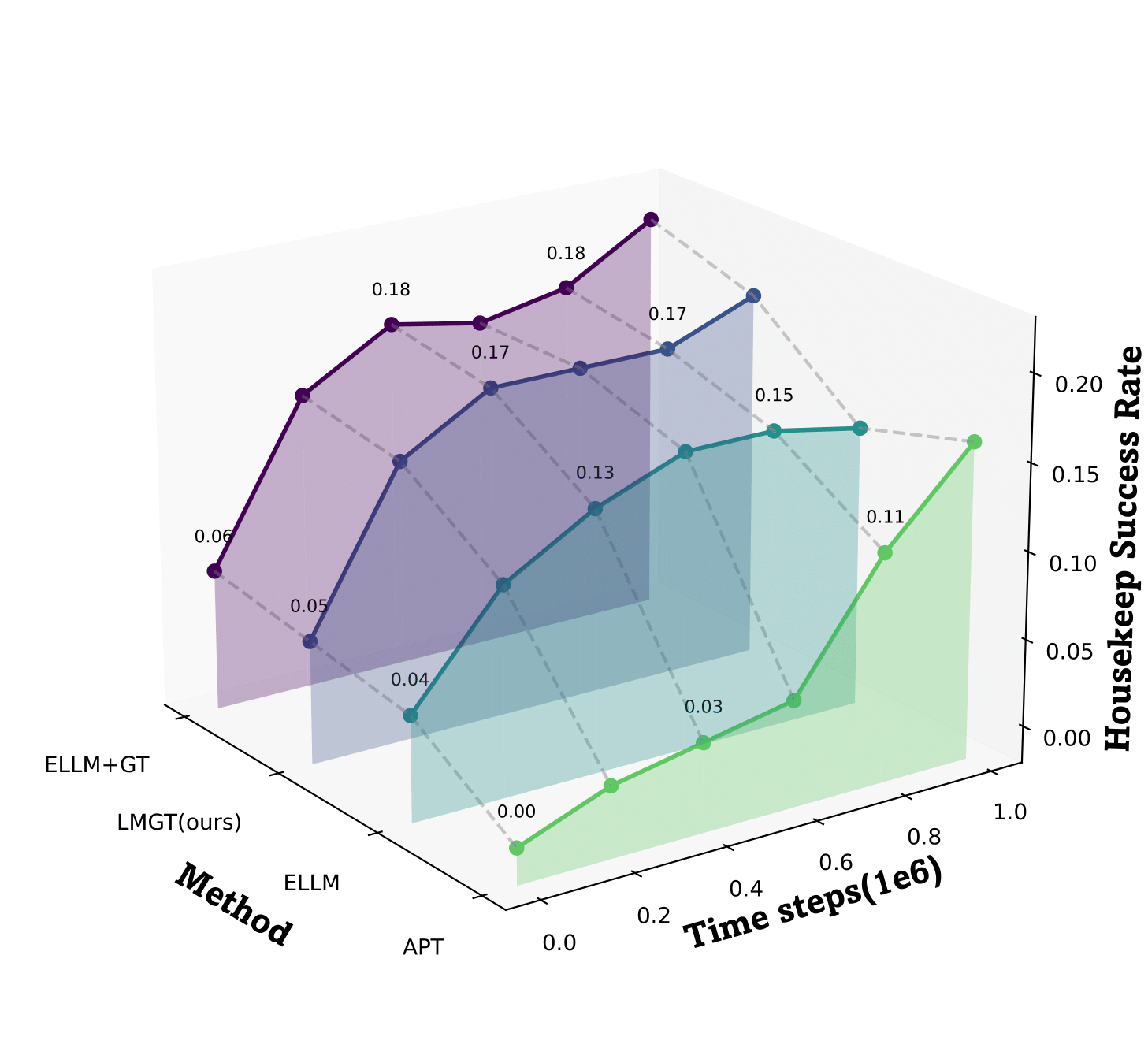}}\hfill
    \subfloat[Task4]{\includegraphics[width=0.25\textwidth]{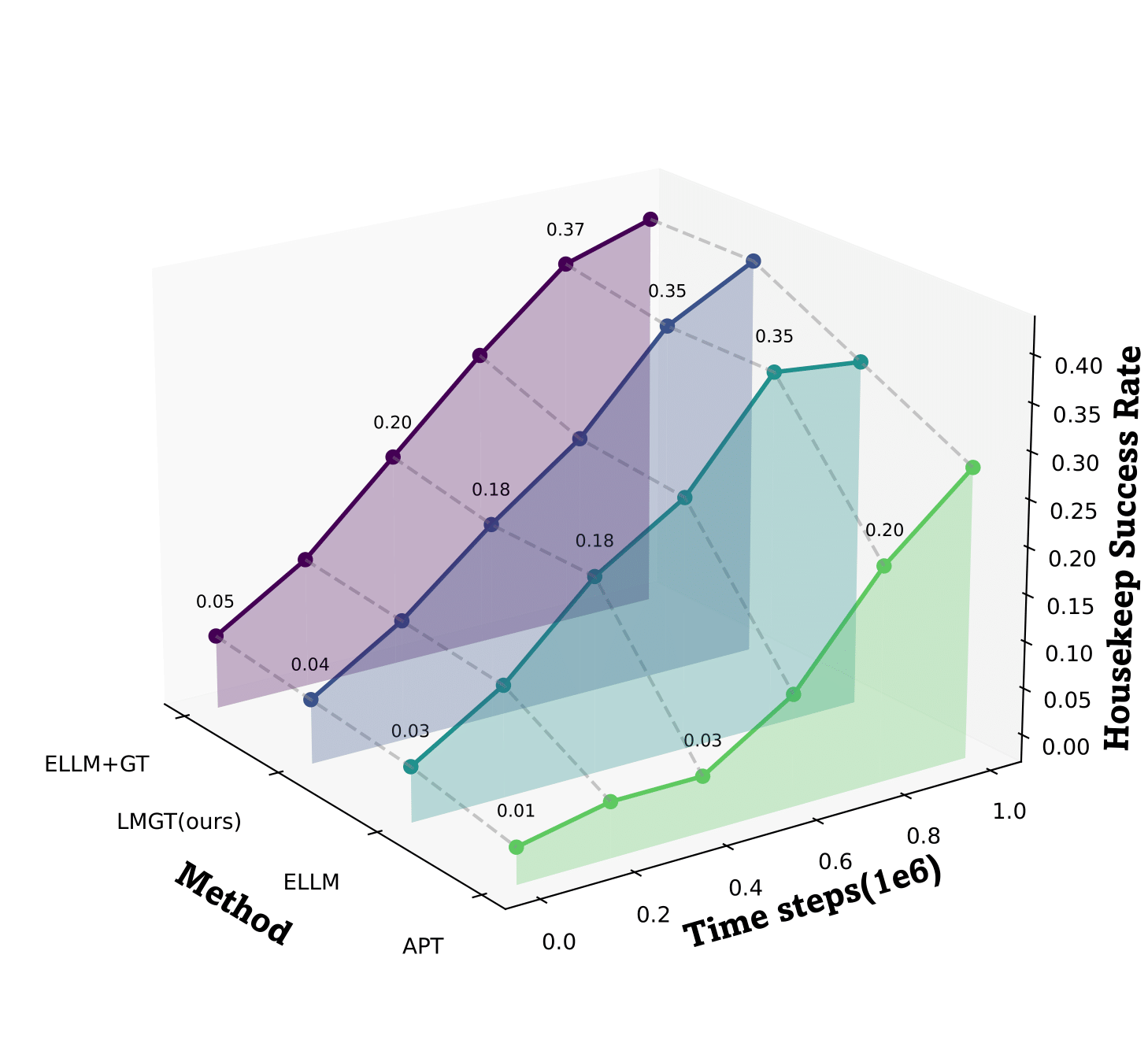}}

    \caption{\textbf{Results from comparative experiments conducted within the Housekeep environment.} Correct arrangement success rates on 4 object-receptacle task sets. }  

    \label{fig:housekeep}
\end{figure*}

We also conducted experiments to assess the performance of different LLMs serving as the ``evaluators'' within our framework, thereby partially evaluating their inferential capabilities, we opted for the Blackjack environment for testing. The experimental results are presented in Table \ref{tab:llm_comparison}. ``Vicuna-30B-4bit-GPTQ'' indicates the use of the Vicuna model, with a size of 30 billion parameters, employing GPTQ quantization with 4-bit precision. ``Llama2-13B-8bit'' signifies the use of the Llama2 model with a size of 13 billion parameters, without any quantization, running in 8-bit floating-point precision. We kept the prompt statements constant by using CoT and Zero-shot prompt, with the inclusion of prior knowledge, and fixed the RL algorithm (A2C).\par

From Table \ref{tab:llm_comparison}, we observe that the precision of quantization has a limited impact on inferential capabilities in the same model. A well-considered quantization method can effectively mitigate the performance loss resulting from quantization. Model size, on the other hand, has a more significant influence on a model's inferential capabilities, a minimally sized language model fails to yield any significant improvement. Additionally, models of identical scale exhibit variations in their inferential capabilities, confined solely within the scope of our framework.\par

\subsubsection{Ablation study} 

\begin{table}[htbp]
\centering
\caption{\textbf{Ablation studies in the Blackjack environment.} Performance comparison of different algorithms under varying state formats and training durations. Positive values indicate improvement, while negative values (in \textcolor{red}{red}) indicate performance degradation.}
\label{tab:ablation_blackjack}
\begin{adjustbox}{width=0.48\textwidth}
\begin{tabular}{lccccccc}
\toprule
& & \multicolumn{6}{c}{\textbf{Observable Environmental State Format}} \\
\cmidrule{3-8}
& & \multicolumn{3}{c}{Box} & \multicolumn{3}{c}{Human} \\
\cmidrule(lr){3-5} \cmidrule(l){6-8}
\textbf{Algorithm} & & \multicolumn{1}{c}{n=100} & \multicolumn{1}{c}{n=1000} & \multicolumn{1}{c}{n=10000} & \multicolumn{1}{c}{n=100} & \multicolumn{1}{c}{n=1000} & \multicolumn{1}{c}{n=10000} \\
\midrule
DQN & & 0.08 & 0.03 & 0.18 & 0.01 & 0.00 & \textcolor{red}{-0.01} \\
PPO & & 0.05 & 0.20 & 0.15 & \textcolor{red}{-0.01} & 0.12 & \textcolor{red}{-0.10} \\
A2C & & 0.32 & 0.12 & 0.13 & 0.01 & \textcolor{red}{-0.03} & \textcolor{red}{-0.01} \\
\bottomrule
\end{tabular}
\end{adjustbox}
\end{table}

In Section \ref{main exp}, we noted that requiring a LLM to perform multiple tasks simultaneously within a single query might compromise its capability \cite{clavie2023large}. Based on this principle, we designed an ablation experiment to test the performance of LMGT in both the `box' and `human' formats within a more visually complex Blackjack environment. For the latter, recognizing card information and converting it into numerical data constitutes a highly specialized task. When the LLM must first process complex visual data, its reasoning ability diminishes. The experimental results, as shown in Table \ref{tab:ablation_blackjack}, reveal that LMGT's performance in the `human' format fluctuates around the baseline, indicating performance deterioration in this context. This finding demonstrates that our LMGT effectively leverages the LLM's capabilities to guide the agent's learning: when the LLM's capability is insufficient to provide guidance, the agent's performance reverts to the baseline.

\subsection{Verification Experiments in the Housekeep Environment}
\label{sec:housekeep}

To assess the efficacy of LMGT in complex real-world robotic tasks, we conducted experiments using the Housekeep environment \cite{kant2022housekeep}. Housekeep is an embodied agent simulation where the robot must organize a household by correctly placing misplaced objects into their appropriate containers. The agent must deduce the correct object-container pairings without explicit instructions, based on mappings derived from crowdsourced data on typical object-container associations.
We compared LMGT with the baseline method Active Pre-Training (APT) \cite{liu2021behavior}. APT learns behaviors and representations through active exploration of novel states in reward-free environments, utilizing non-parametric entropy maximization in an abstract representation space. This approach circumvents the challenges of density modeling, enabling better scalability in environments with high-dimensional observations, such as those with image inputs.
To demonstrate the significance of Visual Instruction Tuning (illustrated in Figure \ref{fig:llava}), we introduced ELLM \cite{du2023guiding} as an additional baseline. We evaluated ELLM under two conditions: a) Using ground truth simulator states as LLM input, representing the theoretical optimal performance achievable through textual environmental descriptions; b) Using learned descriptors to characterize the environment, exemplifying the practical performance of captioner-based methodologies.

For consistency, we adopted the experimental setup from ELLM, focusing on a simplified Housekeep subset comprising four distinct scenes. Each scene contained one room with five misplaced objects and various potential containers.

Figure \ref{fig:housekeep} presents the comparative results of LMGT and the baseline methods in the Housekeep environment. LMGT consistently outperformed APT, highlighting the capacity of LLMs to effectively guide agent learning in complex scenarios. Notably, LMGT's performance surpassed that of ELLM when the latter employed learned descriptors. \textbf{Despite similar limitations in accessing ground truth environmental states, LMGT, leveraging Visual Instruction Tuning, retained more information conducive to LLM decision-making}. Moreover, despite ELLM utilizing ground truth values in certain settings, our LMGT was able to match, and in some scenarios even surpass, ELLM's performance without access to ground truth information.

\subsection{Experiments in Industrial Recommendation Scenarios}

In this section, we further apply our framework to Google's RL recommendation algorithm, SlateQ \cite{ie2019slateq}, to elucidate its potential in industrial applications. 
\subsubsection{Simulation environment}
RecSim\cite{ie2019recsim} is a simulation platform for constructing and evaluating recommendation systems that naturally support sequential interactions with users. Developed by Google, it simulates users and environments to assess the effectiveness and performance of recommendation algorithms. We employ RecSim to create an environment that reflects user behavior and item structure to evaluate our LMGT framework.

We construct a ``Choc vs. Kale'' recommendation scenario, where the goal is to maximize user satisfaction and engagement over the long term by recommending a certain proportion of ``chocolate'' and ``kale'' elements. In this scenario, the ``chocolate'' element represents content that is interesting but not conducive to long-term satisfaction, while the ``kale'' element represents relatively less exciting but beneficial content for long-term satisfaction. The recommendation algorithm needs to balance these two elements to achieve maximized long-term user satisfaction. \par

\begin{table}[htbp]
\centering
\caption{\textbf{Impact of different LLMs on framework performance.} Results across various time steps for the Blackjack environment, with higher values indicating better performance.}
\label{tab:llm_comparison}
\small
\begin{adjustbox}{width=0.48\textwidth}
\begin{tabular}{llrrr}
\toprule
\textbf{Model Family} & \textbf{Configuration} & \multicolumn{3}{c}{\textbf{Time Steps}} \\
\cmidrule(lr){3-5}
& & n=100 & n=1000 & n=10000 \\
\midrule
\multirow{9}{*}{Vicuna} 
& 7B-4bit & -0.20 & 0.18 & 0.32 \\
& 7B-8bit & -0.20 & 0.18 & 0.32 \\
& 7B-16bit & -0.20 & 0.18 & 0.32 \\
& 7B-4bit-GPTQ & -0.20 & 0.18 & 0.32 \\
\cmidrule(lr){2-5}
& 13B-4bit & -0.20 & 0.18 & 0.32 \\
& 13B-8bit & 0.10 & 0.18 & 0.34 \\
& 13B-16bit & 0.10 & 0.18 & 0.36 \\
& 13B-4bit-GPTQ & 0.10 & 0.18 & 0.34 \\
\cmidrule(lr){2-5}
& 30B-4bit-GPTQ & \textbf{0.12} & \textbf{0.30} & \textbf{0.45} \\
\midrule
\multirow{8}{*}{Llama2} 
& 7B-4bit & -0.30 & 0.16 & 0.32 \\
& 7B-8bit & -0.30 & 0.16 & 0.32 \\
& 7B-16bit & -0.30 & 0.16 & 0.32 \\
& 7B-4bit-GPTQ & -0.30 & 0.16 & 0.32 \\
\cmidrule(lr){2-5}
& 13B-4bit & 0.10 & 0.16 & 0.32 \\
& 13B-8bit & 0.10 & 0.16 & 0.32 \\
& 13B-16bit & 0.12 & 0.16 & 0.34 \\
& 13B-4bit-GPTQ & 0.12 & 0.16 & 0.34 \\
\bottomrule
\multicolumn{5}{p{0.95\columnwidth}}{\footnotesize Best results are highlighted in bold. All experiments conducted in the Blackjack environment.}\\
\end{tabular}
\end{adjustbox}
\end{table}

\begin{table}[htbp]
\centering
\caption{\textbf{Performance comparison of LMGT framework and baseline SlateQ in recommendation tasks.} Results show average rewards across different training episodes, with LMGT consistently outperforming the baseline method.}
\label{tab:slateq_comparison}
\begin{adjustbox}{width=0.46\textwidth}
\begin{tabular}{llrrr}
\toprule
\multirow{2}{*}{\textbf{Method}} & \multirow{2}{*}{\textbf{Metric}} & \multicolumn{3}{c}{\textbf{Number of Episodes}} \\
\cmidrule(lr){3-5}
& & n=10 & n=50 & n=5000 \\
\midrule
SlateQ & A.R. & 831.082 & 913.528 & 1127.136 \\
\textbf{LMGT (ours)} & A.R. & \textbf{933.624} & \textbf{1125.171} & \textbf{1150.251} \\
\midrule
\textbf{Perf. Gain} & Improvement & +102.542 & +211.643 & +23.115 \\
\bottomrule
\multicolumn{5}{p{0.95\columnwidth}}{\footnotesize A.R.: Average Reward, Perf. Gain: Performance Gain.}\\
\end{tabular}
\end{adjustbox}
\end{table}

In our scenario, the entire simulation environment consists primarily of document models and user models. The document model serves as the main interface for interaction between users and the recommendation system (agent) and is responsible for selecting a subset of documents from a database containing a large number of documents to deliver to the recommendation system. The user model simulates user behavior and reacts to the slates provided by the recommendation system.\par

The database in the document model essentially serves as a container for observable and unobservable features of underlying documents. In this scenario, document attributes are modeled as continuous features with values in the range of $[0,1]$, referred to as the Kaleness scale. A document assigned a score of 0 represents pure ``chocolate'', which is intriguing but regrettable, whereas a document with a score of 1 represents pure``kale'', which is less exciting but nutritious. Additionally, each document has a unique integer ID, and the document model selects $N$ candidate documents in sequential order based on their IDs.\par

The user model includes both observable and unobservable user features. Based on these features, the model responds to the received slate according to certain rules. Each user is characterized by the features of net kale exposure ($nke_{t}$) and satisfaction ($sat_{t}$), which are associated through the sigmoid function $\sigma$ to ensure that $sat_{t}$ is constrained within a bounded range. Specifically, the satisfaction level is modeled as a sigmoid function of the net kale exposure, which determines the user's satisfaction with the recommended slate:\par

\begin{equation}
    sat_{t} = \sigma (\tau \cdot nke_{t})
\end{equation}\par

Where $\tau$ is a user-specific sensitivity parameter. Upon receiving a Slate from the recommendation system, users select items to consume based on the Kaleness scale of the documents. Specifically, for item $i$, the probability of it being chosen is determined by $p \sim e^{1-kaleness(i)}$. After making their selections, the net kale exposure evolves as follows:

\begin{equation}
    nke_{t+1} = \beta \cdot nke_{t} + 2(k_{i}-1/2) + \mathcal{N}(0,\eta)
\end{equation}\par

Where $\beta$ represents a user-specific memory discount, while $k_{i}$ corresponds to the kaleness of the selected item, and $\eta$ denotes some noise standard deviation. Lastly, our focus will be on the user's engagement $s_{i}$, i.e. a log-normal distribution with parameters linearly interpolating between the pure kale response $(\mu_{k}, \sigma_{k})$ and the pure choc response $(\mu_{c},\sigma_{c})$:\par
\begin{equation}
    s_{i} \sim log\mathcal{N}(k_{i}\mu_{k}+(1-k_{i})\mu_{c},k_{i}\sigma_{k}+(1-k_{i})\sigma_{c})
\end{equation}
The satisfaction variable $sat_{t}$ represents the sole dynamic component of the user's state, and thus, we generate the user's observable state based on it. In the simulation, user satisfaction is modeled and computed as a latent state. However, to simulate real-world scenarios, we map the latent state to an observable state by introducing noise to account for user uncertainty.\par
\subsubsection{Experimental results}
The experimental configurations for LMGT and the baseline SlateQ approach are identical. We independently trained agents using both our method and the baseline SlateQ, evaluating their performance over an equivalent number of episodes. In the ``Choc vs. Kale'' scenario, each episode consists of a set number of time steps. As illustrated in Table \ref{tab:slateq_comparison}, our results conclusively show that our approach significantly accelerates skill acquisition in agents, enabling them to adeptly navigate the complex challenges of the environment. This rapid development of expertise leverages prior knowledge and skillfully balances the tension between exploration and exploitation. As a consequence, there is an efficient use of sample resources, leading to a marked decrease in the training costs associated with RL models. \textbf{Nonetheless, our study has its limitations.} A notable omission is the analysis of computational resources required for integrating LLMs into the training process. Future research will focus on optimizing the use of computational resources in RL training by applying prior knowledge while addressing the heightened resource demand that comes with incorporating LLMs. Additionally, we have not yet formulated a theoretical framework to explain how LLMs dynamically influence reward structures. Addressing this represents a promising avenue for future research.

\section{Conclusion}

We introduce LMGT, a novel framework that harnesses the extensive knowledge and sophisticated capabilities of LLMs to enhance RL. LMGT effectively balances exploration and exploitation in RL by leveraging LLMs' domain expertise and information processing abilities, seamlessly integrating with existing RL workflows. Our comprehensive experiments across diverse settings and algorithms demonstrate LMGT's efficacy in optimizing the exploration-exploitation trade-off while simultaneously reducing training costs. To validate its practical applicability, we successfully implemented LMGT in the complex Housekeep robotic environment, highlighting its potential for real-world embodied AI applications. While our study yields promising results, it also opens avenues for future research. Specifically, we need to investigate the computational impact of integrating LLMs into the RL training process. Future work will focus on striking a balance between the enhanced capabilities offered by LLMs and the optimization of computational resource allocation in RL training environments.

\section{Limitations and Future Work}

Despite the promising results demonstrated by LMGT across various environments, our research is not without limitations. This section discusses these constraints and outlines potential directions for future research.
\subsection{Current Limitations}

Our study has several notable limitations. First, we have not comprehensively analyzed the computational overhead introduced by integrating LLMs into the RL training process. While LMGT demonstrates improved sample efficiency, the computational resources required for LLM inference during training represent a significant consideration for practical deployments. This trade-off between enhanced learning performance and computational cost warrants further investigation.

Second, the efficacy of LMGT is inherently dependent on the quality and relevance of prior knowledge embedded within the LLM. In domains where LLMs possess insufficient or inaccurate knowledge, the guidance provided may be suboptimal or potentially misleading. Our experiments in the Blackjack environment with the `human' format (Table \ref{tab:ablation_blackjack}) illustrate this limitation, where performance deteriorated when the LLM was tasked with both processing complex visual information and providing guidance simultaneously.

Third, our current implementation does not fully address the theoretical foundations explaining how LLMs dynamically influence reward structures over time. While we observe empirical improvements, a more rigorous theoretical framework would enhance our understanding of LMGT's underlying mechanisms.

Finally, while we have evaluated LMGT across various environments, our testing in highly complex, dynamic, and partially observable environments remains limited. The scalability of our approach to such scenarios requires further validation.
\subsection{Future Directions}
\subsubsection{Extending LMGT to More Complex Situations}

We envision several promising approaches to extend LMGT to more complex situations. For tasks with high complexity, implementing a hierarchical structure where top-level LLMs provide strategic guidance while lower-level LLMs offer tactical reward shifts could effectively manage complexity through task abstraction at different levels. In scenarios with computational constraints, knowledge distillation from larger LLMs into smaller, specialized models could enable efficient deployment while maintaining guidance quality.

Complex real-world environments often involve diverse sensory inputs. Enhancing LMGT with multimodal fusion mechanisms that coherently integrate visual, audio, and proprioceptive information before generating reward guidance would address this challenge. This integration could be complemented by an adaptive mechanism that dynamically adjusts the magnitude and frequency of reward shifts based on the agent's learning progress and task complexity, optimizing the balance between LLM guidance and autonomous exploration.

For rapidly changing complex environments, extending LMGT with meta-learning capabilities would enable the framework to quickly adapt its reward guidance strategy to new situations based on limited experience. Additionally, in environments with partial observability, incorporating uncertainty estimation in LMGT's reward shifts would provide more conservative guidance when environmental understanding is limited.

For extremely complex scenarios, a hybrid approach where human experts periodically review and refine the LLM's reward guidance could create a continuous improvement cycle, combining the scalability of automated systems with human judgment.
\subsubsection{Theoretical Advancements}

Future research should focus on developing a theoretical framework that explains how LLM-guided reward shifts influence the convergence properties and exploration patterns of RL algorithms. This would include analyzing the relationship between reward shifts and value function initialization, as well as studying how different types of prior knowledge affect exploration-exploitation dynamics.
\subsubsection{Computational Efficiency}

Investigating methods to reduce the computational overhead of LMGT through techniques such as model compression, selective inference, and asynchronous guidance represents another important research direction. Developing strategies that balance the benefits of LLM guidance with computational efficiency would enhance the practical applicability of our framework.
\subsubsection{Multi-Agent and Collaborative Settings}

Extending LMGT to multi-agent reinforcement learning scenarios, where LLMs could provide guidance on collaboration strategies or mediate between competing objectives, offers exciting possibilities. Research in this direction could lead to more effective solutions for complex multi-agent problems in areas such as autonomous vehicle coordination, distributed robotics, and strategic games.
\subsubsection{Personalized and Contextual Guidance}

Developing mechanisms for LMGT to provide personalized guidance based on the specific characteristics and learning history of individual agents could further enhance sample efficiency. This would involve adapting the reward shifts based on the agent's strengths, weaknesses, and learning progress, creating a more tailored learning experience.

By addressing these limitations and pursuing these future directions, we believe LMGT can evolve into a more robust, efficient, and widely applicable framework for enhancing reinforcement learning across diverse domains and applications.

\section{Acknowledgement}
This work is supported by the National Natural Science Foundation of China (62102241), the Shanghai Municipal Natural Science Foundation (23\-ZR\-1425400).

\appendix
\section{Prompt demo}
\label{appendix1}
\begin{figure*}[htbp]
  \centering
  \includegraphics[scale=0.65]{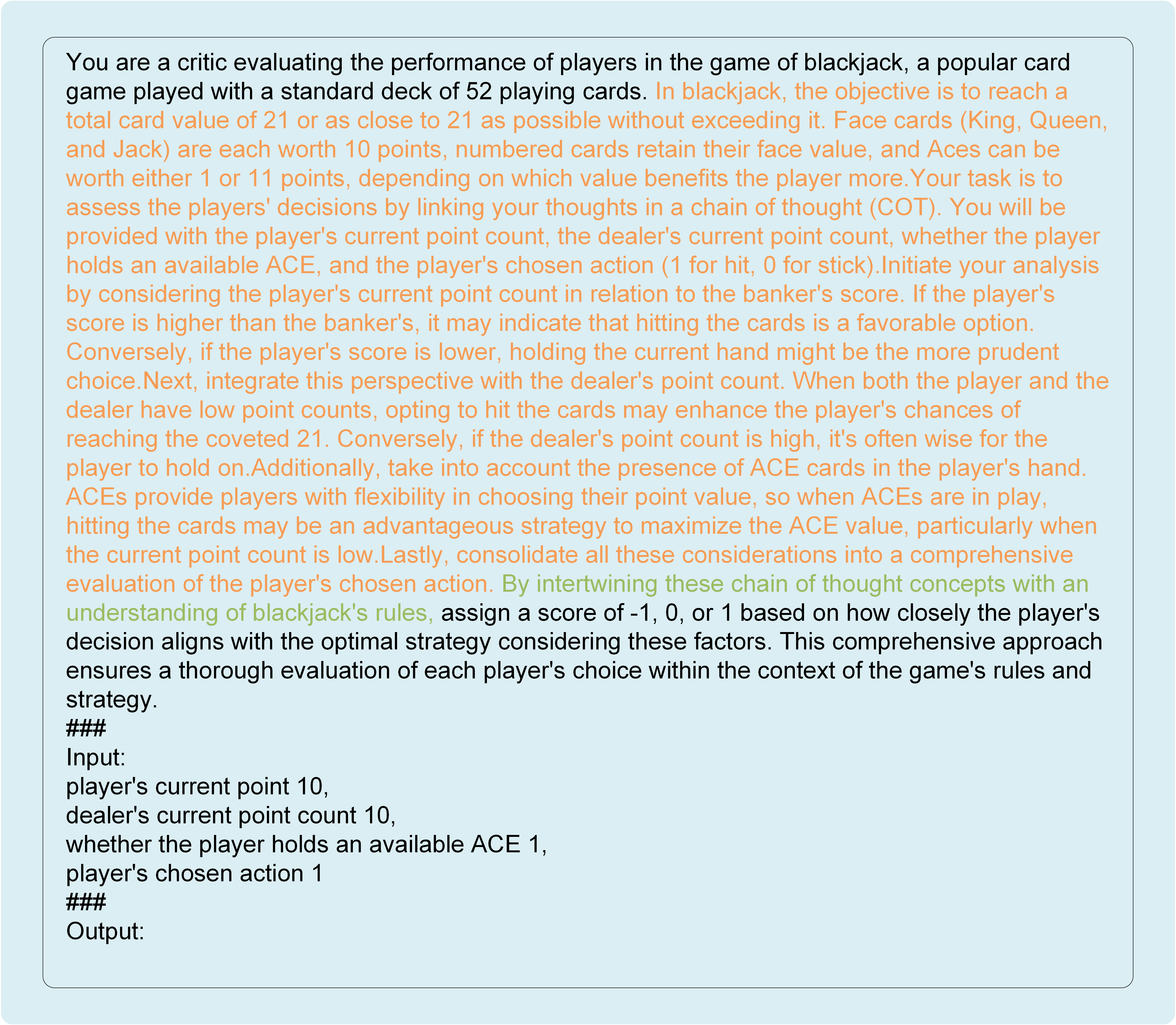}
  \caption{A salient example of a ``prior-knowledge-inclusive prompt statement''.}
  \label{fig3}
\end{figure*}

\begin{figure*}[htbp]
  \centering
  \includegraphics[scale=0.65]{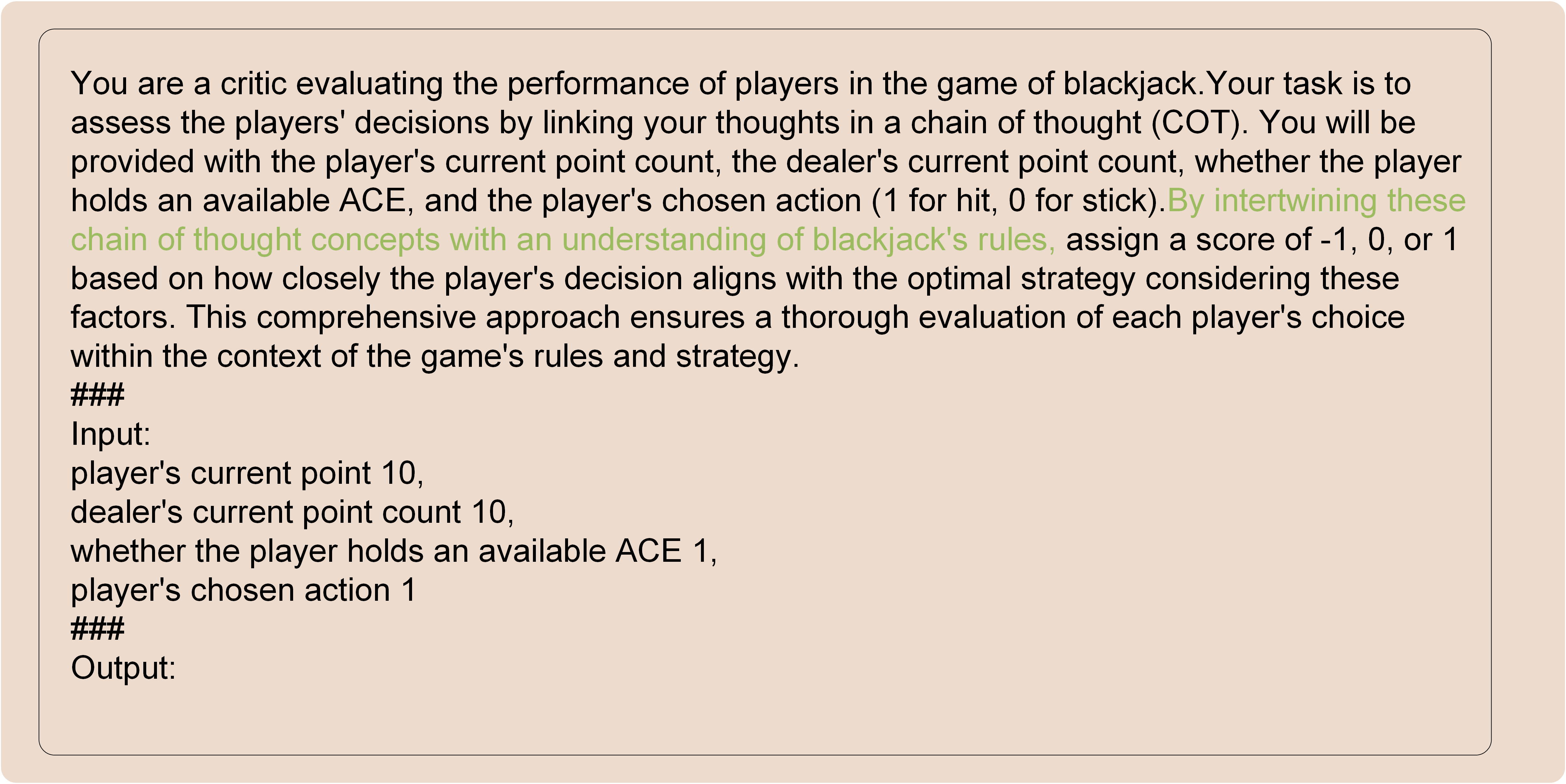}
  \caption{A salient example of a ``prior-knowledge-exclusive prompt statement''.}
  \label{fig4}
\end{figure*}

\begin{figure*}[htbp]
  \centering
  \includegraphics[scale=0.65]{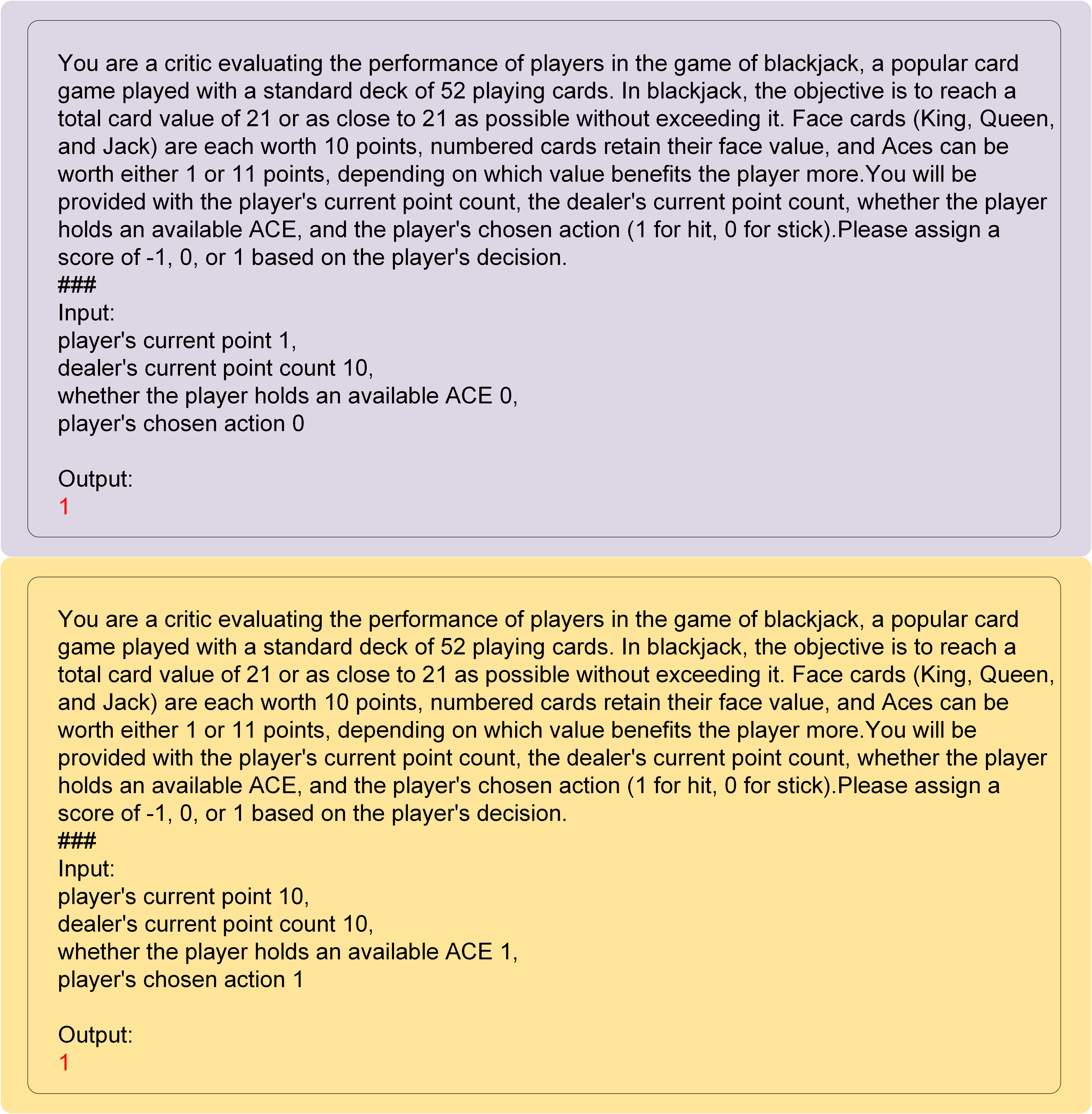}
  \caption{If the model is not required to engage in chain thinking, it will tend to simply provide a fixed result.}
  \label{fig5}
\end{figure*}

\begin{figure*}[htbp]
  \centering
  \includegraphics[scale=0.65]{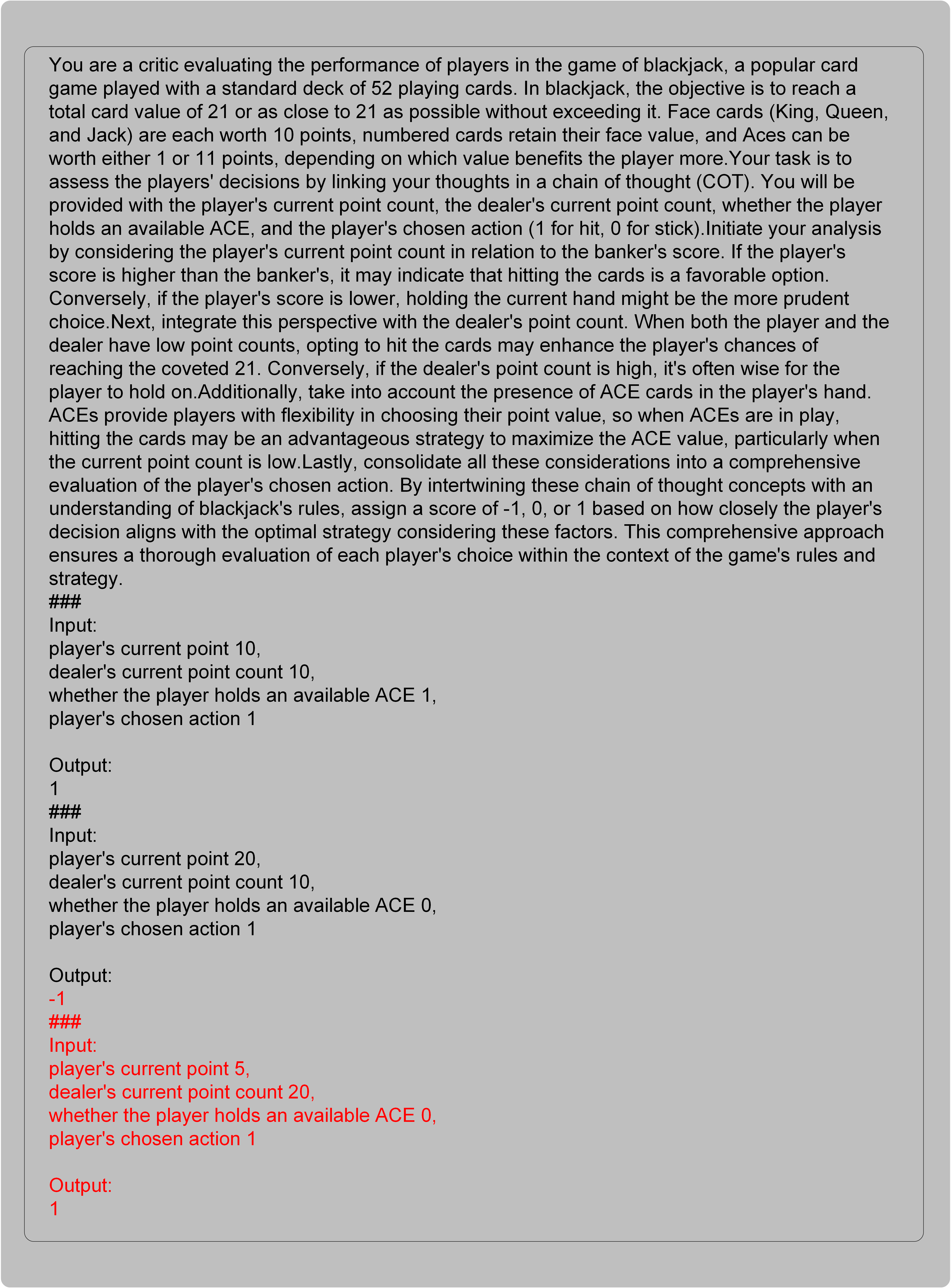}
  \caption{After providing a few examples, the LLM produced some results with hallucinatory elements.}
  \label{fig6}
\end{figure*}

\begin{figure*}[htbp]
  \centering
  \includegraphics[scale=0.65]{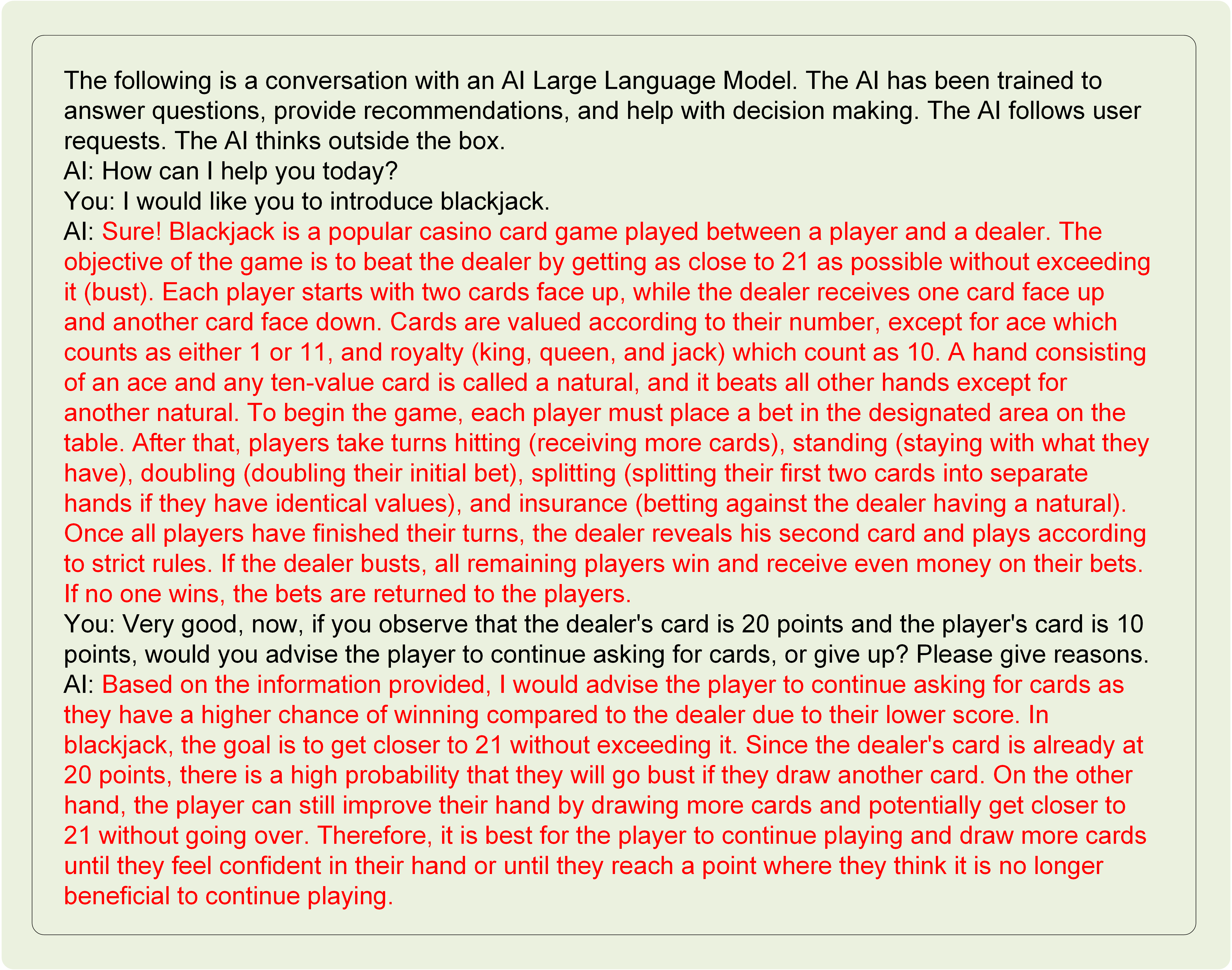}
  \caption{The weights in the model already encompass knowledge about the game of Blackjack.}
  \label{fig7}
\end{figure*}

We provide several prompt statements for demonstration.Figures \ref{fig3} and \ref{fig4} depict the prompt statement designs for ``incorporating prior knowledge'' and ``excluding prior knowledge'', respectively. In Figure \ref{fig5}, it is shown that when LLMs are not required to engage in CoT, they often neglect the provided prompt and produce a consistent output regardless of the circumstances. Figure \ref{fig6} exhibits that when LLMs are given a complete answer example in the prompt, they tend to generate nonsensical content by imitating the example. Figure \ref{fig7} illustrates the inquiries made to the Vicuna-30B model, with the generated results (highlighted in red) indicating that the weights of the Vicuna-30B model include significant prior knowledge related to Blackjack.


\bibliographystyle{elsarticle-num}
\bibliography{ref_new}

\end{document}